\documentclass{article}

\usepackage[numbers]{natbib}
\usepackage[preprint]{neurips_2026}

\usepackage[utf8]{inputenc} 
\usepackage[T1]{fontenc}    
\usepackage{hyperref}       
\usepackage{url}            
\usepackage{booktabs}       
\usepackage{amsfonts}       
\usepackage{nicefrac}       
\usepackage{enumitem}
\usepackage{xcolor}
\usepackage{microtype}      
\usepackage{xcolor}         
\usepackage{graphicx}
\usepackage{amsmath}
\usepackage{wrapfig}  
 \usepackage{algorithm}
 \usepackage{algpseudocode}
 \usepackage{algorithmicx}
\usepackage{subfigure}
\usepackage{float}
\usepackage{amsthm}

\newcommand{\posc}[1]{\textcolor{green!50!black}{#1}}
\newcommand{\negc}[1]{\textcolor{red!70!black}{#1}}
\usepackage{pifont}
\usepackage{booktabs, multirow, xcolor, colortbl, makecell}
\definecolor{improvebg}{RGB}{220,240,220}   
\definecolor{worsenbd}{RGB}{250,220,220}    
\definecolor{promotionbg}{RGB}{240,240,240} 

\title{Does Text Actually Help? Uncovering and Resolving Text Collapse in Multimodal Time Series Forecasting}

\author{%
  Huu Hiep Nguyen \quad Minh Hoang Nguyen \quad Dung Nguyen \quad Hung Le \\
  Applied Artificial Intelligence Initiative \\
Deakin University \\
Geelong, Australia \\
}

\begin{document}

\maketitle

\begin{abstract}
Multimodal time series forecasting, which pairs numerical sequences with domain-relevant textual reports, promises to inject world knowledge into forecasting pipelines. However, we uncover a critical failure mode in existing frameworks that we term text collapse: the text branch converges to a content-independent transformation, contributing negligible discriminative signal regardless of the input description. We argue that text collapse is a consequence of a fundamental asymmetry in time series forecasting: the numerical input is strongly autocorrelated with the output, making the numerical backbone inherently dominant, while the text branch, despite carrying complementary and often critical information, is insufficiently utilized, leading to its systematic underexploitation. To address this, we propose \textbf{REST-TS} (\textbf{R}esidual-\textbf{E}xclusive \textbf{S}upervision for \textbf{T}ext in \textbf{T}ime \textbf{S}eries), which turns the asymmetry into a design principle: the numerical backbone produces its own independent numerical forecast, and the text branch is exclusively supervised to predict the structured components of the residual, the prediction gap that numbers cannot explain. Because no numerical pathway can reduce these losses, the text branch must extract genuine content from the input description. Evaluated across diverse real-world domains and backbone architectures, REST-TS achieves state-of-the-art performance and consistently demonstrates greater text-branch utilization than existing frameworks, providing strong empirical evidence that supervising the text branch on the residual compels it to extract genuine content from the input.

\end{abstract}

\section{Introduction}
\label{sec:intro}

Forecasting future values of a time series is rarely a purely numerical problem. A central banker predicting inflation, an epidemiologist modelling disease spread, or an energy analyst anticipating grid demand all rely heavily on textual context such as policy statements, outbreak bulletins, and weather advisories that numerical history alone cannot convey \citep{10.1145/3711896.3736567,wang2024from}. Recent benchmarks formalise this intuition at scale by pairing historical time series with expert-curated textual reports across diverse real-world domains~\citep{liu2024timemmd}, and a growing family of multimodal forecasting frameworks reports consistent improvements over numerical-only baselines when text is incorporated~\citep{liu2024timemmd, li2026language, nguyen2026spectral}, suggesting that text carries a predictive signal. We find that existing frameworks fail to capture this signal in practice.

\textbf{Does text actually help?} We put this question to a direct test. Rather than evaluating held-out accuracy alone, we probe whether the text branch of each model actually produces representations that vary with the input description, by measuring the effective rank~\citep{roy2007effective} of its output representations across a diverse set of backbone architectures. Effective rank measures how spread the output representations are across inputs: a high value indicates that the encoder produces diverse outputs that vary with the input, while a low value indicates that it maps most inputs to nearly to nearly a single direction regardless of content (formal definition in Appendix~\ref{app:metrics}). As shown in Figure~\ref{fig:effective_rank}, MMTSF (MM-TSFLib~\citep{liu2024timemmd}) and TaTS~\citep{li2026language} both exhibit strikingly low effective rank across all six backbones, revealing that their text encoders produce nearly identical outputs regardless of the input content. The reported multimodal gains thus stem from the model architecture rather than from the textual information itself.

\begin{wrapfigure}{r}{0.40\textwidth}
  \vspace{-6pt}
  \centering
  \includegraphics[width=0.39\textwidth]{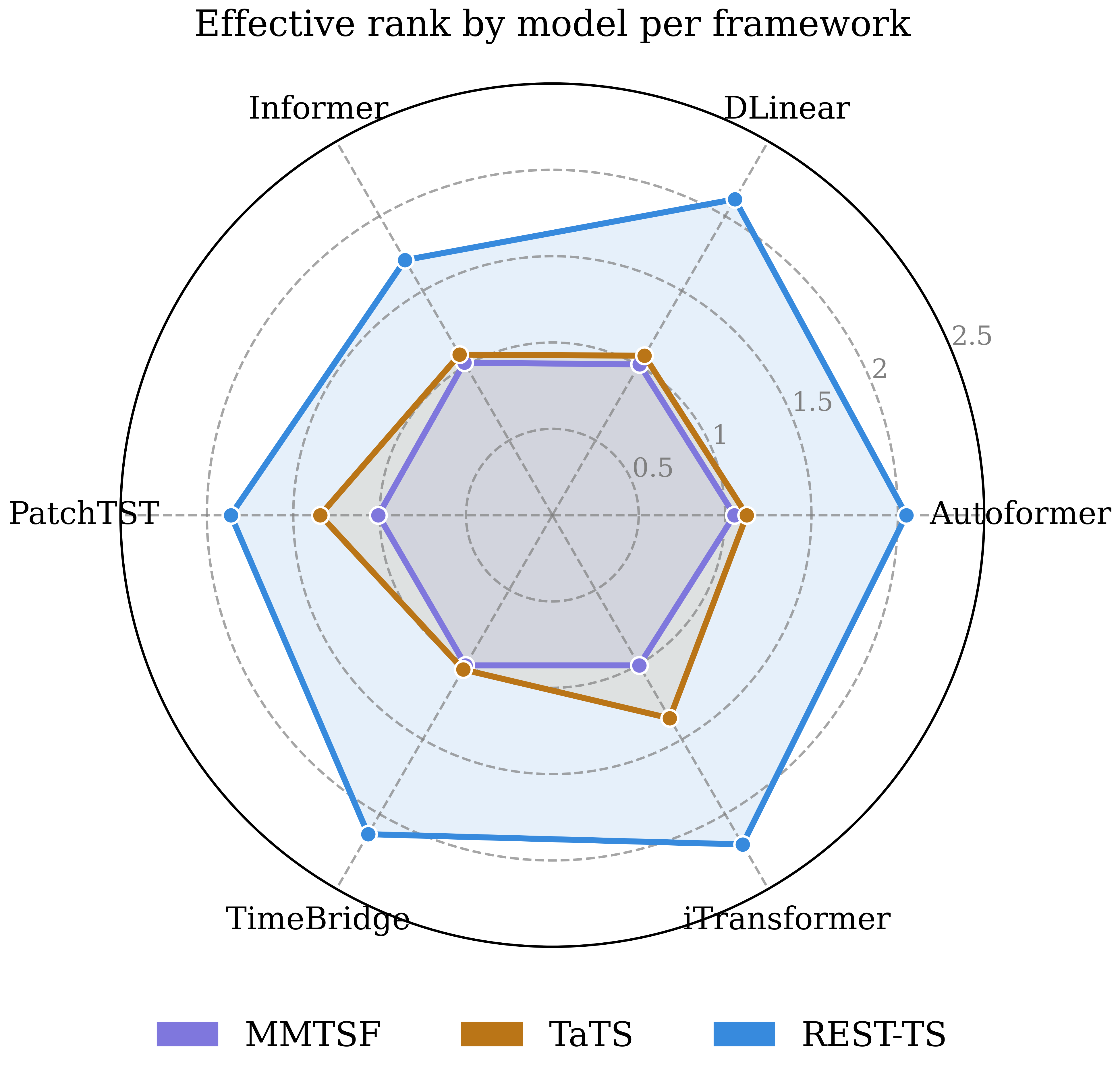}
    \caption{Effective rank of the text branch per backbone.}
    \label{fig:effective_rank}
  \vspace{-4pt}
\end{wrapfigure}

We term this failure mode \textbf{text collapse}: the text branch of a multimodal forecasting model degenerates such that it produces nearly the same output regardless of the input text. Text collapse is not a data problem: we observe it consistently across all domains of Time-MMD~\citep{liu2024timemmd}, a benchmark that pairs time series with high-quality, expert-curated reports, ruling out poor data quality as the cause. Rather, we argue it is rooted in a characteristic that is fundamental to the \emph{nature of multimodal time series forecasting}: the strong autocorrelation between the numerical input $\mathbf{X}$ and the numerical output $\mathbf{Y}$. A backbone trained on $\mathbf{X}$ alone already achieves low forecasting error; the historical series is simply a strong predictor of its own future. By contrast, text provides context for rare, non-periodic, and structurally absent events in numerical history. This inherent asymmetry makes the numerical modality \emph{dominant} in gradient competition: its learning signal is consistently stronger and easier to exploit than text, regardless of the fusion mechanism used. Unless the text branch is given a task that the numerical pathway \emph{cannot} solve, the optimiser will find the numerical shortcut, leading to degeneration and causing the text encoder to produce nearly identical representations regardless of the input description (Figure~\ref{fig:effective_rank}).

To resolve text collapse, we propose \textbf{REST-TS} (\textbf{R}esidual-\textbf{E}xclusive \textbf{S}upervision for \textbf{T}ext in \textbf{T}ime \textbf{S}eries). Rather than competing over the same forecast target, REST-TS assigns text and numbers complementary roles: the numerical backbone produces its own independent numerical forecast, and the text branch is supervised to predict the trend and event components of the residual, the difference between the future target and the numerical forecast. A Trend-Noise-Event decomposition isolates these text-attributable components from noise, and a slowly updated copy of the numerical backbone ensures the resulting targets remain stable and non-trivial throughout training. Because no numerical pathway can reduce these losses, the text branch must attend to the input description. As shown in Figure~\ref{fig:effective_rank},  REST-TS achieves substantially higher effective rank across all six backbones, demonstrating that supervising the text branch on the residual compels it to extract genuine content from the input. In summary, our contributions are:
\begin{itemize}
    \item \textbf{Text collapse identification.} We identify and formally define text collapse, a previously unreported failure mode in multimodal time series forecasting, and demonstrate its prevalence across MM-TSFlib and TaTS through effective rank analysis.

    \item \textbf{REST-TS.} We propose a framework that structurally resolves text collapse by assigning the text branch exclusive, supervised responsibility over the text-attributable components of the backbone residual, enforced by an EMA target network and a Trend-Noise-Event decomposition module.

    \item \textbf{State-of-the-art results.} Experiments across diverse real-world domains and backbone architectures demonstrate that REST-TS consistently outperforms existing multimodal frameworks without requiring architectural modifications to the backbone.
\end{itemize}

\section{Related Work}
\label{sec:related}
\subsection{Unimodal Time Series Forecasting}
\label{sec:related_unimodal}
Deep learning for time series forecasting has progressed through several paradigm shifts. Early Transformer-based models such as Informer~\citep{haoyietal-informer-2021-Informer}, Autoformer~\citep{wu2021autoformer}, and FEDformer~\citep{zhou2022fedformer} tailored self-attention to long-horizon dependencies. \citet{Zeng_Chen_Zhang_Xu_2023} then showed that a simple linear model with moving-average decomposition (DLinear) could match or outperform these Transformers, highlighting decomposition as a key predictive ingredient. Subsequent work rehabilitated Transformers through better tokenization and attention design, including PatchTST~\citep{nie2023a-patchtst}, iTransformer~\citep{liu2024itransformer}, TimeMixer~\citep{wang2023timemixer}, and TimeBridge~\citep{liu2025timebridge}. Although these methods perform strongly, they are unable to leverage external information from other modalities, such as text, which provides rich semantic context for time series analysis.
\subsection{Multimodal Time Series Forecasting}
\label{sec:related_multimodal}

The Time-MMD benchmark~\citep{liu2024timemmd} and its companion library MM-TSFlib represent the first systematic framework for studying text-augmented time series forecasting at scale, combining any numerical backbone with a frozen LLM text encoder and fusing the two predictions through a learned linear combination. TaTS~\citep{li2026language} encodes paired texts into temporal representations that mirror the periodic structure of the series and injects them at the embedding level. SpecTF~\citep{nguyen2026spectral} projects both modalities into spectral space and fuses them via cross-attention that reweights frequency bands based on textual relevance.
Despite this diversity of fusion strategies, none of the above works examine whether the text branch genuinely extracts content-dependent signals from paired reports, a question our paper asks directly. We conduct the first diagnostic study of text collapse in multimodal time series forecasting and demonstrate that existing fusion designs provide no structural incentive for the text encoder to attend to input content. REST-TS addresses this by assigning the text branch exclusive, supervised prediction targets that the numerical branch cannot absorb, enforcing content-dependent text utilization by construction.

\subsection{Modality Collapse in Multimodal Learning}
\label{sec:related_collapse}
The failure of one modality to contribute meaningfully in a jointly trained multimodal model has been extensively studied in vision-language and audio-visual settings under terms including modality imbalance, greedy learning, and modality under-utilization. Early work documented that multimodal classifiers often underperform unimodal counterparts and proposed gradient rebalancing as a remedy~\citep{wang2020gradient, peng2022ogmge}. \citet{huang2021makes, huang2022modality} provide theoretical characterization of multimodal learning, proving that late-fusion networks under SGD can deterministically suppress weaker modalities. More recent remedies include PMR~\citep{fan2023pmr}, MMPareto~\citep{wei2024mmpareto}, MLA~\citep{zhang2024mla}, and CGGM~\citep{guo2024classifierguided}, which tackle the problem via gradient manipulation, Pareto optimization, alternating unimodal adaptation, and classifier-guided modulation, respectively. Despite this diversity of fusion strategies, none of the above works examine whether the text branch genuinely extracts content-dependent signals from paired reports, a question our paper asks directly. We conduct the first diagnostic study of text collapse in multimodal time series forecasting and demonstrate that existing fusion designs \emph{permit} content-dependent text utilization but do not \emph{enforce} it, leaving the text encoder free to learn shortcuts that bypass input content. REST-TS addresses this by assigning the text branch exclusive, supervised prediction targets that the numerical branch cannot absorb, enforcing content-dependent text utilization by construction.

\section{Method}
\label{sec:method}

\begin{figure}[t]
    \centering
    \includegraphics[width=\linewidth]{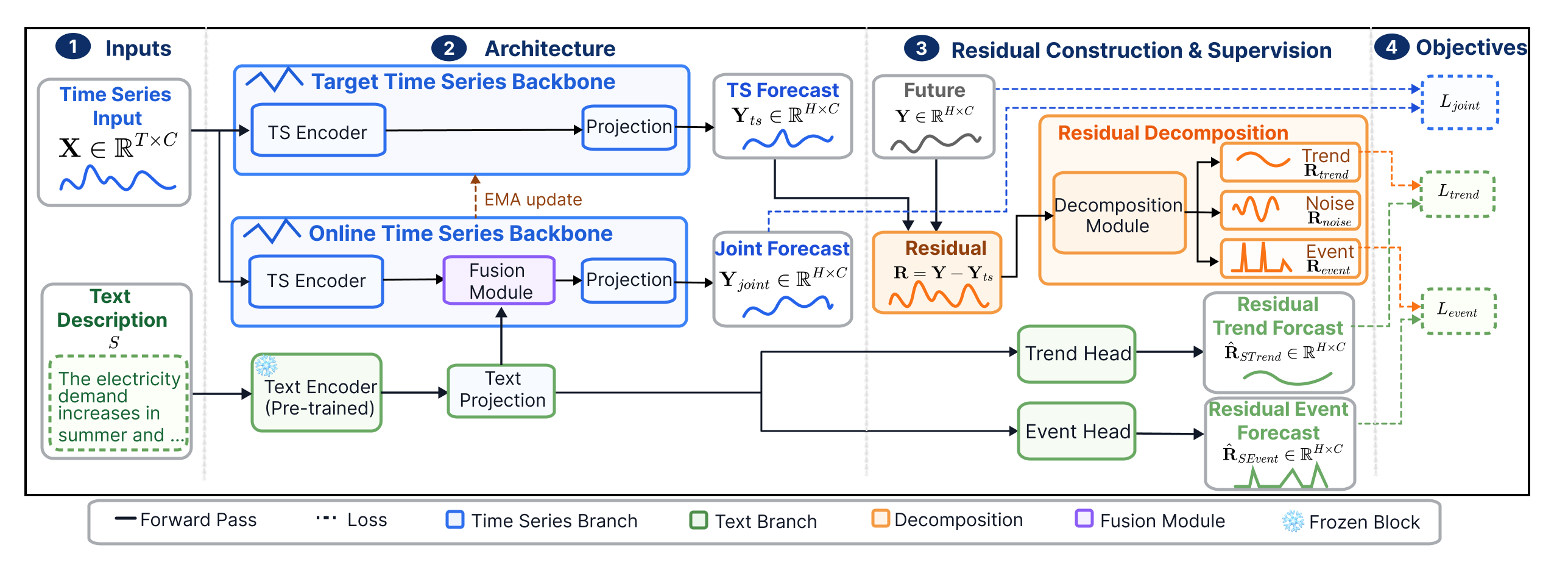}
    \caption{ \textbf{Overview of REST-TS.} The \textbf{Online Backbone} (bottom, blue) is trained with backpropagation and fuses its numerical latent with the text representation $\mathbf{Z}_{S}$ to produce the joint forecast $\mathbf{Y}_{\mathrm{joint}}$. The \textbf{Target Network} (top, blue) is a momentum-updated EMA copy that produces a stable unimodal forecast $\mathbf{Y}_{\mathrm{ts}}$, defining the residual $\mathbf{R} = \mathbf{Y} - \mathbf{Y}_{\mathrm{ts}}$. The \textbf{Text Branch} (green) encodes $S$ with a frozen language model and feeds the result into the Fusion Module and two prediction heads. $\mathbf{R}$ is decomposed (orange) into trend, noise, and event components; only $\mathbf{R}_{\mathrm{trend}}$ and $\mathbf{R}_{\mathrm{event}}$ supervise the text heads exclusively, forcing the text encoder to capture contextual signals that historical numbers alone are insufficient to predict.
}
    \label{fig:pipeline}
\end{figure}

\subsection{Problem Formulation}
\label{sec:problem}

Let $\mathbf{X} \in \mathbb{R}^{T \times C}$ denote a multivariate time series of length $T$ over $C$ variates, and let $S$ denote a natural-language text description that summarises domain-level information relevant to the observed period. The multimodal forecasting task is to predict the future horizon $\mathbf{Y} \in \mathbb{R}^{H \times C}$, where $H$ is the forecast horizon, by jointly leveraging $\mathbf{X}$ and $S$.

As demonstrated in Section~\ref{sec:intro}, existing late-output fusion frameworks suffer from \emph{text collapse}: the text encoder maps every input to nearly the same representation. REST-TS resolves this by assigning the text branch an \emph{exclusive, supervised prediction target}, the variance that the time-series-only prediction leaves behind. The following subsections describe each component in detail.

\subsection{Overall Architecture}
\label{sec:overview}
 

Figure~\ref{fig:pipeline} illustrates REST-TS. The \emph{time series branch} (Section~\ref{sec:ts_branch}) produces both the joint multimodal forecast $\mathbf{Y}_{\mathrm{joint}}$ and a gradient-free EMA forecast $\mathbf{Y}_{\mathrm{ts}}$, the latter defining the residual $\mathbf{R} = \mathbf{Y} - \mathbf{Y}_{\mathrm{ts}}$. The \emph{text branch} (Section~\ref{sec:text_branch}) encodes the description $S$ with a frozen language model and feeds two prediction heads for the trend and event components of $\mathbf{R}$. The \emph{Trend--Noise--Event decomposition} (Section~\ref{sec:decomp}) splits $\mathbf{R}$ into three components, of which only the trend and event components supervise the text heads, a target that the numerical stream \emph{cannot} produce, structurally preventing text collapse without any gradient surgery. 

\subsection{Time Series Branch}
\label{sec:ts_branch}

\paragraph{Multimodal TS forecast.}
The numerical input $\mathbf{X} \in \mathbb{R}^{T \times C}$ is processed by a trainable online backbone $\mathcal{F}_\theta$, instantiated as any standard time series backbone (e.g., PatchTST~\cite{nie2023a-patchtst}, iTransformer~\cite{liu2024itransformer}, TimeMixer~\cite{wang2023timemixer}), to produce a latent representation:
\begin{equation}
    \mathbf{Z}_{\mathrm{ts}} = \mathcal{F}_\theta(\mathbf{X}) \in \mathbb{R}^{D \times C},
    \label{eq:encoder}
\end{equation}
where $D$ is the latent dimension and $\theta$ denotes the online parameters updated via backpropagation. REST-TS imposes no architectural constraint on $\mathcal{F}_\theta$, making the framework backbone-agnostic. To produce a text-informed forecast, $\mathbf{Z}_{\mathrm{ts}}$ is fused with the text representation $\mathbf{Z}_{S}$ (Section~\ref{sec:text_branch}) through a cross-modal fusion module:
\begin{equation}
    \mathbf{Z}_{\mathrm{joint}} = \mathrm{Fusion}(\mathbf{Z}_{\mathrm{ts}},\, \mathbf{Z}_{S}) \in \mathbb{R}^{D \times C},
    \label{eq:fusion}
\end{equation}
implemented by default as element-wise addition, though any fusion mechanism can be substituted. The fused representation is then projected to the forecast horizon by a trainable projection layer $\mathrm{Proj}_{\phi}$:
\begin{equation}
    \mathbf{Y}_{\mathrm{joint}} = \mathrm{Proj}_{\phi}(\mathbf{Z}_{\mathrm{joint}}) \in \mathbb{R}^{H \times C}.
    \label{eq:yjoint}
\end{equation}
At inference, $\mathbf{Y}_{\mathrm{joint}}$ is the final forecast; all other components introduced below exist solely to shape the training signal.

\paragraph{Unimodal TS forecast.}
A central design principle of REST-TS is that the unimodal forecast $\mathbf{Y}_{\mathrm{ts}}$, which defines the residual decomposition targets for the text branch, must be stable and gradient-free. To achieve this, we maintain a target network $(\mathcal{F}_{\theta'}, \mathrm{Proj}_{\phi'})$ consisting of separate parameter copies of both the online backbone $\mathcal{F}_\theta$ and the online projection $\mathrm{Proj}_\phi$, updated solely via exponential moving average (EMA) after each training step:
\begin{equation}
    (\theta', \phi') \;\leftarrow\; m\,(\theta', \phi') + (1 - m)\,(\theta, \phi),
    \label{eq:ema}
\end{equation}
where $m \in (0,1)$ is the momentum coefficient. The target network is never updated by backpropagation and produces the unimodal forecast:
\begin{equation}
    \mathbf{Y}_{\mathrm{ts}} = \mathrm{Proj}_{\phi'}\!\left(\mathcal{F}_{\theta'}(\mathbf{X})\right) \in \mathbb{R}^{H \times C}.
    \label{eq:yts}
\end{equation}
Decoupling $\mathbf{Y}_{\mathrm{ts}}$ from the online gradient flow serves two purposes. First, the EMA update acts as a low-pass filter on the targets, letting $\mathbf{R}_{\mathrm{trend}}$ and $\mathbf{R}_{\mathrm{event}}$ evolve smoothly rather than forcing the text heads to chase a non-stationary objective. Second, because $\theta'$ lags $\theta$, the residual $\mathbf{R} = \mathbf{Y} - \mathbf{Y}_{\mathrm{ts}}$ stays larger than the online backbone's in-sample residual; EMA does not prevent residual shrinkage, but slows it enough to give the text branch a sustained window of non-trivial supervision.

\subsection{Text Branch}
\label{sec:text_branch}

\paragraph{Text encoder.}
The text description $S$ is encoded by a pre-trained language model
whose weights are kept \emph{frozen} throughout training:
\begin{equation}
    \mathbf{Z}_{S}^{\mathrm{enc}} = \mathrm{TextEnc}(S) \in \mathbb{R}^{D_{enc}},
    \label{eq:textenc}
\end{equation}
where $D_{enc}$ is the dimension of TextEnc output. Freezing the encoder preserves pre-trained linguistic knowledge and
prevents gradient interference from $\mathcal{L}_{\mathrm{joint}}$.
The encoded representation is then passed through a trainable linear
projection $\mathrm{TextProj}$ to match the model's latent dimension:
\begin{equation}
    \mathbf{Z}_{S} = \mathrm{TextProj}(
        \mathbf{Z}_{S}^{\mathrm{enc}}) \in \mathbb{R}^{D \times C}.
    \label{eq:textproj}
\end{equation}


\paragraph{Residual prediction heads.}
A core observation motivating REST-TS is that linguistic descriptions of time series domains encode two qualitatively different types of information: slow-moving \emph{trend signals} (e.g., seasonal demand patterns, long-run growth) and abrupt \emph{event signals} (e.g., policy shocks, market disruptions, extreme weather events). We instantiate two lightweight MLP heads operating on $\mathbf{Z}_{S}$ to predict
each:
\begin{align}
    \hat{\mathbf{R}}_{S,\mathrm{trend}} &= \mathrm{TrendHead}(\mathbf{Z}_{S})
        \in \mathbb{R}^{H \times C}, \label{eq:trend_head} \\
    \hat{\mathbf{R}}_{S,\mathrm{event}} &= \mathrm{EventHead}(\mathbf{Z}_{S})
        \in \mathbb{R}^{H \times C}. \label{eq:event_head}
\end{align}
Each head is an MLP with GeLU activation. Their supervision targets: $\mathbf{R}_{\mathrm{trend}}$ and $\mathbf{R}_{\mathrm{event}}$, are derived from the decomposition stream (Section~\ref{sec:decomp}). This exclusivity is the primary guarantee against text collapse: the text encoder \emph{must} extract content-dependent signals from $S$ to minimise the residual supervision losses, as no numerical pathway exists to do so.

\subsection{Trend--Noise--Event Decomposition}
\label{sec:decomp}

The residual $\mathbf{R} = \mathbf{Y} - \mathbf{Y}_{\mathrm{ts}}$ captures the prediction gap between the ground truth and the time-series-only forecast $\mathbf{Y}_{\mathrm{ts}}$. The central assumption motivating our decomposition is that \emph{not all of this gap is attributable to text}: some portion arises from stochastic noise that no textual description can predict, while the rest reflects genuine contextual signals, such as trends and event-driven shifts, that are described in $S$. Supervising the text branch on the raw residual would therefore corrupt the learning signal with unpredictable noise, making it harder for the text encoder to extract meaningful content. We verify this empirically in our ablation: removing the decomposition consistently degrades performance across both backbones (Table~\ref{tab:ablation_components}). We instead decompose $\mathbf{R}$ into three components according to their textual attributability:

\begin{itemize}
    \item \textbf{Trend} ($\mathbf{R}_{\mathrm{trend}}$): smooth, slow-moving deviations driven by gradual contextual shifts described in $S$, such as seasonal demand patterns or long-run growth trends;
    \item \textbf{Noise} ($\mathbf{R}_{\mathrm{noise}}$): high-frequency stochastic oscillations that arise from random fluctuations and carry no textual signal;
    \item \textbf{Event} ($\mathbf{R}_{\mathrm{event}}$): sharp, non-periodic spikes caused by exogenous shocks explicitly described in $S$, such as sudden policy changes, market disruptions, or extreme weather events.
\end{itemize}
Under this assumption, only $\mathbf{R}_{\mathrm{trend}}$ and $\mathbf{R}_{\mathrm{event}}$ carry extractable textual signal and should supervise the text branch. Supervising on $\mathbf{R}_{\mathrm{noise}}$ would force the text encoder to predict variance that is by definition unpredictable from $S$, injecting harmful gradient noise and undermining content-dependent learning.

To extract these components, we apply a Trend--Noise--Event decomposition. Let $\mathrm{MAvg}$ denote a moving-average filter and let $\mathrm{TopK}_{\mathrm{FFT}}(\cdot,\,K)$ denote the operation that reconstructs a signal using only its $K$ highest-frequency Fourier components. The decomposition proceeds as:
\begin{align}
    \mathbf{R}_{\mathrm{trend}}  &= \mathrm{MAvg}(\mathbf{R}),
        \label{eq:rtrend} \\
    \mathbf{R}_{\mathrm{noise}}  &= \mathrm{TopK}_{\mathrm{FFT}}\!\left(
        \mathbf{R} - \mathbf{R}_{\mathrm{trend}},\; K\right),
        \label{eq:rnoise} \\
    \mathbf{R}_{\mathrm{event}}  &= \mathbf{R}
        - \mathbf{R}_{\mathrm{trend}}
        - \mathbf{R}_{\mathrm{noise}}.
        \label{eq:revent}
\end{align}

The trend component $\mathbf{R}_{\mathrm{trend}}$ captures the slow, smooth deviation driven by gradual contextual shifts described in $S$. The noise component $\mathbf{R}_{\mathrm{noise}}$ isolates the $K$ highest-frequency oscillations in the de-trended residual, corresponding to stochastic fluctuations that carry no textual signal. The event component $\mathbf{R}_{\mathrm{event}}$ is the remainder after removing both trend and noise: it retains non-periodic, high-magnitude spikes caused by exogenous shocks explicitly described in $S$, such as sudden policy changes, market disruptions, or extreme weather events. The hyperparameter $K$ controls the noise bandwidth; lowering $K$ attributes more high-frequency energy to the event component, while raising $K$ attributes more to the noise component. The noise component $\mathbf{R}_{\mathrm{noise}}$ is \emph{not} supervised by any text head: high-frequency stochastic oscillations carry no extractable textual signal. Only $\mathbf{R}_{\mathrm{trend}}$ and $\mathbf{R}_{\mathrm{event}}$ serve as exclusive supervision targets for the text branch, corresponding to the two types of temporal phenomena most reliably described in domain-level textual reports.

\subsection{Training Objective}
\label{sec:loss}

REST-TS is trained end-to-end with a composite loss consisting of three terms. The \emph{joint forecast loss} penalises errors in the primary multimodal prediction:
\begin{equation}
    \mathcal{L}_{\mathrm{joint}} = \mathrm{MSE}(\mathbf{Y}_{\mathrm{joint}},\,
    \mathbf{Y}).
    \label{eq:ljoint}
\end{equation}
The \emph{trend} and \emph{event supervision losses} enforce content-dependent text utilisation. We apply the stop-gradient operator $\mathrm{sg}(\cdot)$~\citep{chen2021simsiam} to the decomposition targets:
\begin{align}
    \mathcal{L}_{\mathrm{trend}} &= \mathrm{MSE}\!\left(
        \hat{\mathbf{R}}_{S,\mathrm{trend}},\;
        \mathrm{sg}\!\left(\mathbf{R}_{\mathrm{trend}}\right)
    \right), \label{eq:ltrend} \\
    \mathcal{L}_{\mathrm{event}} &= \mathrm{MSE}\!\left(
        \hat{\mathbf{R}}_{S,\mathrm{event}},\;
        \mathrm{sg}\!\left(\mathbf{R}_{\mathrm{event}}\right)
    \right). \label{eq:levent}
\end{align}
The $\mathrm{sg}(\cdot)$ operator treats $\mathbf{R}_{\mathrm{trend}}$ and $\mathbf{R}_{\mathrm{event}}$ as constants during backpropagation, so that gradients from $\mathcal{L}_{\mathrm{trend}}$ and $\mathcal{L}_{\mathrm{event}}$ flow only through the text heads $\hat{\mathbf{R}}_{S,\mathrm{trend}}$ and $\hat{\mathbf{R}}_{S,\mathrm{event}}$. This ensures that the text branch is the sole component responsible for reducing the residual supervision losses, preventing the numerical branch from absorbing the text supervision signal.
The total training objective is:
\begin{equation}
    \mathcal{L} = \mathcal{L}_{\mathrm{joint}}
    + \lambda_1\,\mathcal{L}_{\mathrm{trend}}
    + \lambda_2\,\mathcal{L}_{\mathrm{event}},
    \label{eq:total_loss}
\end{equation}
where $\lambda_1, \lambda_2 \geq 0$ are scalar hyperparameters controlling the relative weight of each supervision signal. At inference time, only $\mathbf{Y}_{\mathrm{joint}}$ is used as the final forecast; the residual heads and the decomposition module are discarded, incurring no additional inference overhead.

\section{Experiments}
\label{sec:experiments}

\begin{table*}[!t]
\centering
\caption{%
  Time-series forecasting performance (MSE) on Time-MMD benchmark. \textbf{Bold} is the best. \textit{Promotion} denotes the percentage MSE reduction of REST-TS compared to the best baseline TaTS. Results are averaged across all prediction lengths; full results across all backbones and prediction lengths are in Appendix~\ref{app:full_forecast_timemmd}.%
}
\label{tab:main_results}
\resizebox{\textwidth}{!}{%
\begin{tabular}{l l ccccccccc}
\toprule
\multicolumn{2}{c}{Method} &
  Agri. & Clim. & Eco. & Ener. & Envi. & Heal. & Sec. & Soc. & Traf. \\
\midrule
\multirow{5}{*}{\rotatebox{90}{TimeBridge}}
& Uni-modal                & 0.117 & 1.555 & 0.0182  & 0.305 & 0.350 & 1.968 & 130.27 & 1.291 & 0.174 \\
& MM-TSFLib                & 0.121 & 1.262 & 0.1014 & 0.286 & 0.320 & 1.504 & 140.32 & 1.314 & 0.213 \\
& TaTS                     & 0.120 & 1.198 & 0.0105 & 0.297 & 0.274 & 1.364 & 126.64 & 1.124 & \textbf{0.160} \\
& \textbf{REST-TS (ours)}  & \textbf{0.099} & \textbf{0.955} & \textbf{0.0083} & \textbf{0.261} & \textbf{0.270} & \textbf{1.252} & \textbf{108.82} & \textbf{0.941} & 0.175 \\
\rowcolor{promotionbg} & Promotion
  & \cellcolor{improvebg}\posc{17.5\%}
  & \cellcolor{improvebg}\posc{20.3\%}
  & \cellcolor{improvebg}\posc{21.0\%}
  & \cellcolor{improvebg}\posc{12.1\%}
  & \cellcolor{improvebg}\posc{1.5\%}
  & \cellcolor{improvebg}\posc{8.2\%}
  & \cellcolor{improvebg}\posc{14.1\%}
  & \cellcolor{improvebg}\posc{16.3\%}
  & \cellcolor{worsenbd}\negc{$-$9.4\%} \\
\midrule
\multirow{5}{*}{\rotatebox{90}{iTransformer}}
& Uni-modal                & 0.122 & 1.183 & 0.0140 & 0.269 & 0.441 & 1.587 & 115.94 & 1.212 & 0.213 \\
& MM-TSFLib                & 0.119 & 1.049 & 0.1051 & 0.256 & 0.307 & 1.647 & 140.43 & 1.219 & 0.248 \\
& TaTS                     & 0.114 & 1.035 & \textbf{0.0082} & 0.276 & 0.271 & 1.290 & 119.71 & 1.039 & 0.198 \\
& \textbf{REST-TS (ours)}  & \textbf{0.113} & \textbf{1.022} & 0.0084 & \textbf{0.251} & \textbf{0.259} & \textbf{1.275} & \textbf{109.73} & \textbf{0.981} & \textbf{0.182} \\
\rowcolor{promotionbg} & Promotion
  & \cellcolor{improvebg}\posc{0.9\%}
  & \cellcolor{improvebg}\posc{1.3\%}
  & \cellcolor{worsenbd}\negc{$-$2.4\%}
  & \cellcolor{improvebg}\posc{9.1\%}
  & \cellcolor{improvebg}\posc{4.4\%}
  & \cellcolor{improvebg}\posc{1.2\%}
  & \cellcolor{improvebg}\posc{8.3\%}
  & \cellcolor{improvebg}\posc{5.6\%}
  & \cellcolor{improvebg}\posc{8.1\%} \\
\midrule
\multirow{5}{*}{\rotatebox{90}{PatchTST}}
& Uni-modal                & 0.120 & 1.220 & 0.0172  & 0.269 & 0.552 & 1.652 & 112.85 & 1.097 & 0.188 \\
& MM-TSFLib                & 0.115 & 1.016 & 0.1143 & 0.263 & 0.345 & 1.422 & 133.15 & 1.208 & 0.216 \\
& TaTS                     & \textbf{0.114} & 1.019 & \textbf{0.0087} & \textbf{0.258} & 0.287 & 1.357 & 120.52 & 0.990 & 0.180 \\
& \textbf{REST-TS (ours)}  & 0.115 & \textbf{0.995} & 0.0090 & \textbf{0.258} & \textbf{0.272} & \textbf{1.272} & \textbf{109.21} & \textbf{0.974} & \textbf{0.178} \\
\rowcolor{promotionbg} & Promotion
  & \cellcolor{worsenbd}\negc{$-$0.9\%}
  & \cellcolor{improvebg}\posc{2.4\%}
  & \cellcolor{worsenbd}\negc{$-$3.4\%}
  & \cellcolor{improvebg}\posc{0.0\%}
  & \cellcolor{improvebg}\posc{5.2\%}
  & \cellcolor{improvebg}\posc{6.3\%}
  & \cellcolor{improvebg}\posc{9.4\%}
  & \cellcolor{improvebg}\posc{1.6\%}
  & \cellcolor{improvebg}\posc{1.1\%} \\
\midrule
\multirow{5}{*}{\rotatebox{90}{DLinear}}
& Uni-modal                & 0.223 & 1.190 & 0.0580  & 0.291 & 0.558 & 1.737 & 109.11 & 1.151 & 0.230 \\
& MM-TSFLib                & 0.221 & 1.040 & 0.1771 & 0.313 & 0.296 & 1.635 & 112.63 & 1.297 & 0.412 \\
& TaTS                     & 0.229 & 0.939 & 0.0241 & 0.324 & 0.296 & 1.466 & 112.53 & 1.302 & 0.196 \\
& \textbf{REST-TS (ours)}  & \textbf{0.138} & \textbf{0.932} & \textbf{0.0131} & \textbf{0.260} & \textbf{0.271} & \textbf{1.319} & \textbf{107.02} & \textbf{0.867} & \textbf{0.186} \\
\rowcolor{promotionbg} & Promotion
  & \cellcolor{improvebg}\posc{39.7\%}
  & \cellcolor{improvebg}\posc{0.7\%}
  & \cellcolor{improvebg}\posc{45.4\%}
  & \cellcolor{improvebg}\posc{19.8\%}
  & \cellcolor{improvebg}\posc{8.4\%}
  & \cellcolor{improvebg}\posc{10.0\%}
  & \cellcolor{improvebg}\posc{4.9\%}
  & \cellcolor{improvebg}\posc{33.4\%}
  & \cellcolor{improvebg}\posc{5.1\%} \\
\bottomrule
\end{tabular}
}
\end{table*}

\begin{table}[t]
\centering
\small
\setlength{\tabcolsep}{4pt}
\caption{MSE on FNF and FNSPID datasets. \textbf{Bold} is best and \underline{underlined} is second-best per backbone column. Results are averaged across all prediction lengths; full results across all backbones and prediction length are in Appendix~\ref{app:full_forecast_fin}.}
\label{tab:mse_all}
\resizebox{\textwidth}{!}{%
\begin{tabular}{ll cc cc cc}
\toprule
\multirow{2}{*}{} & \multirow{2}{*}{Dataset} 
  & \multicolumn{2}{c}{\textbf{REST-TS (ours)}} 
  & \multicolumn{2}{c}{MMTSF} 
  & \multicolumn{2}{c}{TaTS} \\
\cmidrule(lr){3-4}\cmidrule(lr){5-6}\cmidrule(lr){7-8}
 & & PatchTST & iTransformer & PatchTST & iTransformer 
   & PatchTST & iTransformer \\
\midrule
\multirow{3}{*}{FNF}
  & Bitcoin
    & \textbf{0.005} & \textbf{0.006} & 0.013 & 0.011
    & \underline{0.006} & \textbf{0.006} \\
  & Electricity
    & \textbf{0.177} & \textbf{0.170} & 0.195 & 0.183
    & \underline{0.185} & \underline{0.179} \\
  & Web Traffic
    & \underline{0.681} & \textbf{0.633} & 0.704 & 0.651
    & \textbf{0.675} & \underline{0.644} \\
\midrule
\multirow{6}{*}{FNSPID}
  & Delta (DAL)
    & \textbf{0.018} & \textbf{0.019} & 0.028 & 0.027
    & \textbf{0.018} & \textbf{0.019} \\
  & IBM
    & \textbf{0.161} & \underline{0.170} & 0.283 & 0.336
    & \underline{0.163} & \textbf{0.165} \\
  & JPMorgan (JPM)
    & \textbf{0.097} & \textbf{0.103} & 0.185 & 0.205
    & \underline{0.104} & \underline{0.106} \\
  & NVIDIA (NVDA)
    & \textbf{0.059} & \textbf{0.061} & 0.140 & 0.099
    & \underline{0.061} & \underline{0.063} \\
  & Pfizer (PFE)
    & \textbf{0.045} & \textbf{0.045} & 0.105 & 0.101
    & \underline{0.047} & \textbf{0.045} \\
  & Tesla (TSLA)
    & \textbf{0.726} & \underline{0.866} & 1.414 & 1.686
    & \underline{0.750} & \textbf{0.825} \\
\bottomrule
\end{tabular}
}
\end{table}

\begin{table}[!t]
\centering
\small
\caption{Average MSE comparison of REST-TS vs.\ SpecTF on Time-MMD. \textbf{Bold} is the best. Results are averaged
across all prediction lengths; full results across all backbones and prediction length are in Appendix~\ref{app:full_forecast_spectf}.}
\label{tab:spectf_avg_transposed}
\setlength{\tabcolsep}{4pt}
\resizebox{0.9\columnwidth}{!}{
\begin{tabular}{lrrrrrrrrr}
\toprule
Method & Agri. & Clim. & Eco. & Ener. & Envi. & Heal. & Sec. & Soc. & Traf. \\
\midrule
SpecTF & 0.103 & 0.939 & \textbf{0.0085} & 0.246 & \textbf{0.260} & \textbf{1.276} & 108.41 & 0.962 & 0.171 \\
\textbf{REST-TS (ours)} & \textbf{0.100} & \textbf{0.932} & 0.0092 & \textbf{0.218} & 0.260 & 1.295 & \textbf{107.76} & \textbf{0.913} & \textbf{0.170} \\
\textit{Promotion} & \cellcolor{improvebg}+2.3\% & \cellcolor{improvebg}+0.7\% & \cellcolor{worsenbd}-8.2\% & \cellcolor{improvebg}+11.1\% & \cellcolor{improvebg}0.0\% & \cellcolor{worsenbd}-1.5\% & \cellcolor{improvebg}+0.6\% & \cellcolor{improvebg}+5.1\% & \cellcolor{improvebg}+0.7\% \\
\bottomrule
\end{tabular}}
\end{table}

\subsection{Experimental Setup}
\label{sec:setup}

\paragraph{Datasets.}
We evaluate on \textbf{Time-MMD}~\citep{liu2024timemmd}, which pairs multivariate time series with expert-curated domain reports across nine real-world domains, and two financial benchmarks, \textbf{FNF}~\citep{wang2024from} and \textbf{FNSPID}~\citep{dong2024fnspid}, where price series are paired with news articles. Dataset statistics are in Appendix~\ref{app:data}.

\paragraph{Baselines and backbones.}
We compare against \textbf{Uni-modal} (numerical-only), \textbf{MMTSF}~\citep{liu2024timemmd}, and \textbf{TaTS}~\citep{li2026language}. REST-TS is backbone-agnostic and is evaluated with eight architectures: TimeBridge, iTransformer, PatchTST, Crossformer, DLinear, FEDformer, Autoformer, and Informer. We report MSE averaged over all prediction horizons; full implementation details are in Appendix~\ref{app:impl_setup}.

\subsection{Main Results}
\label{sec:main_results}

Table~\ref{tab:main_results} reports MSE on Time-MMD dataset across four representative backbones, averaged over all prediction horizons. REST-TS consistently outperforms all baselines across the large majority of configurations. Qualitative prediction examples across all nine domains are visualised in Appendix~\ref{app:pred_vis}.

\textbf{Comparison with the unimodal backbone.} REST-TS outperforms the numerical-only Uni-modal baseline in 35 of 36 backbone-domain configurations in Table~\ref{tab:main_results}, with the only exception being a marginal $0.001$ MSE difference for TimeBridge on Traffic. Gains are largest in domains where textual context is most predictive, such as Economy and Climate, where REST-TS with TimeBridge reduces MSE by 53.9\% and 38.6\% respectively. In domains with weaker textual signal such as Energy and Traffic, improvements are more modest but remain consistent across all backbones.
 
\textbf{Comparison with multimodal frameworks.} REST-TS outperforms both MMTSF and TaTS in 32 of 36 backbone-domain configurations. The largest gain (45.4\% on DLinear / Economy) arises because DLinear's linear decomposition cannot capture event-driven economic shocks, but those events are densely described in the paired financial reports, giving the text branch a residual target both large and recoverable from text. The largest regression is $-$9.4\% on TimeBridge / Traffic, where TimeBridge already attains a low MSE on the periodic Traffic series and the paired text appears to provide limited additional signal. Full results across all seven backbones are in Appendix~\ref{app:full_forecast_timemmd}.

\textbf{Generalisation to financial benchmarks.} Table~\ref{tab:mse_all} shows that REST-TS achieves the best MSE on the large majority of FNF and FNSPID series. On FNSPID, REST-TS outperforms MMTSF by an average of 48\% across all six companies with PatchTST, with the largest gains on NVIDIA and Pfizer. On FNF, REST-TS matches or improves over TaTS on five of the six PatchTST/iTransformer columns, confirming that residual-exclusive supervision generalises to news-driven financial text beyond the Time-MMD distribution. Full results across prediction lengths are in Appendix~\ref{app:full_forecast_fin}.

\textbf{Adaptation to Standalone Architectures.} To validate that REST-TS is backbone-agnostic beyond standard Transformer-based backbones, we integrate it with SpecTF~\citep{nguyen2026spectral}, a standalone frequency-domain architecture that fuses text and time series via spectral cross-attention. Replacing SpecTF's original text fusion objective with our exclusive residual supervision improves over the original SpecTF in six of the nine Time-MMD domains (Table~\ref{tab:spectf_avg_transposed}), with the largest gains in Energy ($+$11.1\%) and SocialGood ($+$5.1\%); per-horizon results are in Appendix~\ref{app:full_forecast_spectf}.

\textbf{Comparison with modality-balancing baselines.} We further compare REST-TS against OGM-GE~\citep{peng2022ogmge} on PatchTST and iTransformer. REST-TS outperforms OGM-GE on every domain (Table~\ref{tab:ogm_mse_avg}), demonstrating that residual-exclusive supervision provides advantages beyond modality-balancing approaches alone. Full results and text collapse analysis are in Appendix~\ref{app:ogm_comparison}.

\begin{table}[!t]
\centering
\small
\caption{Average MSE comparison of REST-TS vs.\ OGM-GE on Time-MMD. \textbf{Bold} is the best per column per backbone. Results are averaged across all prediction lengths.}
\label{tab:ogm_mse_avg}
\setlength{\tabcolsep}{4pt}
\resizebox{\textwidth}{!}{
\begin{tabular}{ll rrrrrrrrr}
\toprule
Backbone & Method & Agri. & Clim. & Eco. & Ener. & Envi. & Heal. & Sec. & Soc. & Traf. \\
\midrule
\multirow{3}{*}{PatchTST}
& OGM-GE & 0.165 & 1.004 & 0.120 & 0.320 & 0.346 & 1.406 & 126.48 & 1.184 & 0.291 \\
& \textbf{REST-TS (ours)} & \textbf{0.115} & \textbf{0.995} & \textbf{0.0090} & \textbf{0.258} & \textbf{0.272} & \textbf{1.272} & \textbf{109.21} & \textbf{0.974} & \textbf{0.178} \\
& \textit{Promotion} & \cellcolor{improvebg}+30.5\% & \cellcolor{improvebg}+0.9\% & \cellcolor{improvebg}+92.5\% & \cellcolor{improvebg}+19.3\% & \cellcolor{improvebg}+21.4\% & \cellcolor{improvebg}+9.5\% & \cellcolor{improvebg}+13.7\% & \cellcolor{improvebg}+17.8\% & \cellcolor{improvebg}+38.9\% \\
\midrule
\multirow{3}{*}{iTransformer}
& OGM-GE & 0.179 & 1.028 & 0.087 & 0.313 & 0.308 & 1.534 & 129.12 & 1.210 & 0.427 \\
& \textbf{REST-TS (ours)} & \textbf{0.113} & \textbf{1.022} & \textbf{0.0084} & \textbf{0.251} & \textbf{0.259} & \textbf{1.275} & \textbf{109.73} & \textbf{0.981} & \textbf{0.182} \\
& \textit{Promotion} & \cellcolor{improvebg}+36.8\% & \cellcolor{improvebg}+0.6\% & \cellcolor{improvebg}+90.3\% & \cellcolor{improvebg}+19.8\% & \cellcolor{improvebg}+15.8\% & \cellcolor{improvebg}+16.9\% & \cellcolor{improvebg}+15.0\% & \cellcolor{improvebg}+18.9\% & \cellcolor{improvebg}+57.4\% \\
\bottomrule
\end{tabular}}
\end{table}

\subsection{Analysis}
\label{sec:analysis}

\begin{figure}[t]
    \centering
    \includegraphics[width=\linewidth]{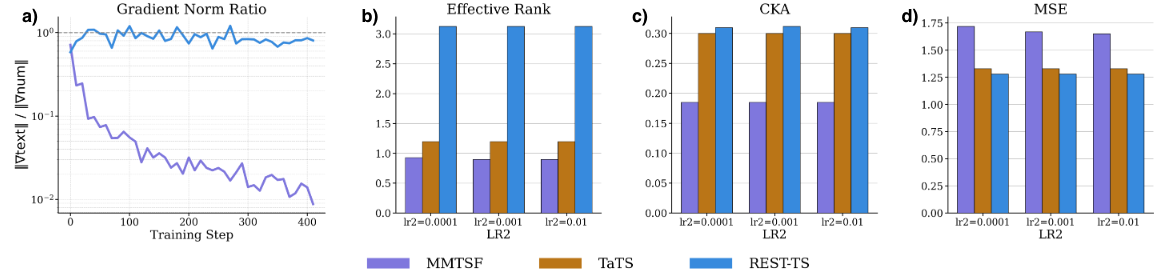}
    \caption{Robustness to text-branch learning rate $\mathrm{lr}_2$. REST-TS (blue) maintains stable effective rank, CKA, and MSE across all settings, while MMTSF (purple) and TaTS (orange) show higher variance, particularly on forecasting error.}
    \label{fig:anal_grad_lr}
\end{figure}

\paragraph{Text gradient dominance.}
A direct signature of text collapse is that the text gradient becomes negligible relative to the numerical gradient during training. Figure~\ref{fig:anal_grad_lr}a plots the ratio $\|\nabla_{\mathrm{text}}\| / \|\nabla_{\mathrm{num}}\|$ over training steps for the Environment domain with iTransformer. MMTSF (purple) decays to $10^{-2}$ within the first 500 steps and never recovers, confirming text collapse. REST-TS (blue) stabilises near parity throughout, a direct consequence of the exclusive residual supervision. Extended analysis across all nine domains and both PatchTST and iTransformer is in Appendix~\ref{app:gradient}.

\paragraph{Robustness to text-branch learning rate.} Figure~\ref{fig:anal_grad_lr}b evaluates effective rank, CKA (formal definitions in Appendix~\ref{app:metrics}), and MSE across three text-branch learning rates on the Health domain with iTransformer. For MMTSF, a higher $\mathrm{lr}_2$ reduces MSE yet effective rank and CKA remain low, revealing that the gain is an optimisation artefact rather than better text utilisation. REST-TS maintains stable effective rank, CKA, and MSE across all three rates, because its text branch is anchored to residual targets that are decoupled from the numerical loss scale. Extended results across all nine domains are in Appendix~\ref{app:lr_sensitivity}.

\paragraph{Prediction divergence.} To further verify that REST-TS genuinely uses text at inference time, we measure the prediction divergence $\delta$: how much the forecast changes when text is removed, normalised by the model's own MSE. A near-zero $\delta$ indicates the prediction is insensitive to the text input, a direct behavioural signature of text collapse. REST-TS consistently achieves the highest $\delta$ across all nine domains and both Informer and PatchTST backbones, while MMTSF and TaTS show near-zero values in several domains, confirming that their predictions are largely unaffected by the text input. Full results are in Appendix~\ref{app:pred_divergence} and per-domain effective-rank and CKA radar charts across six backbones are in Appendix~\ref{app:per_domain}.

\paragraph{Hyperparameter sensitivity.} Sweeping the moving-average kernel width and the Fourier component count $K$ on all nine Time-MMD domains with two backbones (Appendix~\ref{app:hp_sensitivity}) shows that the gap between Informer and TimeBridge is consistently an order of magnitude larger than the variation from either hyperparameter, so backbone choice dominates hyperparameter choice. The small remaining variation is domain-dependent, suggesting $w$ be matched to the timescale of the trend signal and $K$ to the noise level of the series rather than tuned to a single global optimum.
\subsection{Ablation Study}
\label{sec:ablation}

\begin{table}[t]
\vspace{-8pt}
\centering
\caption{Ablation study across three representative domains. Each variant 
is defined by the combination of components: \textit{Residual Decomp.} 
and \textit{Target Network}. \textbf{Bold} = best per column.}
\label{tab:ablation_components}
\resizebox{0.9\textwidth}{!}{%
\begin{tabular}{l cc cc cc cc cc}
\toprule
\multirow{2}{*}{Backbone}
  & \multicolumn{2}{c}{Components}
  & \multicolumn{2}{c}{Climate}
  & \multicolumn{2}{c}{Health}
  & \multicolumn{2}{c}{Environment}
  & \multicolumn{2}{c}{Avg.} \\
\cmidrule(lr){2-3}\cmidrule(lr){4-5}\cmidrule(lr){6-7}
\cmidrule(lr){8-9}\cmidrule(lr){10-11}
  & \makecell{Residual\\Decomp.} & \makecell{Target\\Network}
  & MSE & MAE & MSE & MAE & MSE & MAE & MSE & MAE \\
\midrule
\multirow{4}{*}{PatchTST}
  & \ding{55}        & \ding{55}
    & 1.003 & 0.791 & 1.281 & 0.749 & 0.278 & 0.377 & 0.854 & 0.639 \\
  & $\checkmark$     & \ding{55}
    & 0.996 & \textbf{0.788} & 1.276 & 0.745 & 0.273 & 0.373 & 0.848 & \textbf{0.635} \\
  & \ding{55}        & $\checkmark$
    & 0.998 & 0.789 & 1.277 & 0.748 & 0.275 & 0.374 & 0.850 & 0.637 \\
  & $\checkmark$     & $\checkmark$
    & \textbf{0.995} & \textbf{0.788} & \textbf{1.272} & \textbf{0.744}
    & \textbf{0.272} & \textbf{0.372} & \textbf{0.846} & \textbf{0.635} \\
\midrule
\multirow{4}{*}{iTransformer}
  & \ding{55}        & \ding{55}
    & 1.020 & 0.800 & 1.282 & 0.744 & 0.263 & 0.369 & 0.854 & 0.638 \\
  & $\checkmark$     & \ding{55}
    & 1.008 & 0.799 & 1.276 & \textbf{0.740} & \textbf{0.259} & 0.369
    & 0.852 & 0.635 \\
  & \ding{55}        & $\checkmark$
    & 1.015 & \textbf{0.796} & 1.279 & 0.744 & 0.262 & 0.370
    & 0.852 & 0.635 \\
  & $\checkmark$     & $\checkmark$
    & \textbf{0.995} & 0.797 & \textbf{1.275} & \textbf{0.740}
    & \textbf{0.259} & \textbf{0.361} & \textbf{0.842} & \textbf{0.633} \\
\bottomrule
\end{tabular}
}
\vspace{-4pt}
\end{table}


Table~\ref{tab:ablation_components} ablates Residual Decomposition and the EMA Target Network across three domains with PatchTST and iTransformer. Each component improves over the no-component baseline on its own, and enabling both yields the best average for both backbones, confirming that the two are complementary: Decomposition provides semantically structured targets while the Target Network keeps them informative as the online backbone improves. Synergy is largest in Environment, where event-driven residual structure benefits most from both.

\section{Conclusion}
\label{sec:conclusion}
We identified \emph{text collapse}, a previously unreported failure mode in multimodal time series forecasting in which the text branch degenerates to a content-independent transformation under the autocorrelative dominance of the numerical modality. To resolve it, we proposed REST-TS, which assigns the text branch exclusive supervision over the trend and event components of the backbone residual, decomposed via a Trend--Noise--Event module and stabilised by an EMA target network. Across nine Time-MMD domains and two out-of-domain financial benchmarks, REST-TS consistently outperforms existing multimodal frameworks and achieves higher effective rank and CKA, empirically confirming that the text branch contributes a genuine predictive signal.
\bibliographystyle{plainnat}
\bibliography{ref}

\clearpage
\appendix

\begin{center}
    \Large\bfseries Appendix for \\
    \textit{Does Text Actually Help? Uncovering and Resolving Text Collapse in Multimodal Time Series Forecasting}
\end{center}

\section{Implementation Details}
\label{app:impl}

\subsection{Experimental Setup}
\label{app:impl_setup}

\textbf{Backbone architectures and framework compatibility.} REST-TS is evaluated with eight time series backbones: PatchTST~\citep{nie2023a-patchtst}, iTransformer~\citep{liu2024itransformer}, TimeBridge~\citep{liu2025timebridge}, Autoformer~\citep{wu2021autoformer}, Informer~\citep{haoyietal-informer-2021-Informer}, Corssformer~\citep{zhang2023crossformer}, FEDformer~\citep{zhou2022fedformer} and DLinear~\citep{Zeng_Chen_Zhang_Xu_2023}. Each backbone is used as a drop-in replacement for both the online encoder $\mathcal{F}_\theta$ and the EMA target network $\mathcal{F}_{\theta'}$ without modification to the backbone's internal layers, confirming that REST-TS is backbone-agnostic. The seven Transformer-based backbones reuse their existing forecast head as $\mathrm{Proj}_\phi$; DLinear is the only exception, since it maps directly from $\mathbb{R}^T$ to $\mathbb{R}^H$ with no intermediate latent, and we therefore add a linear projection $\mathrm{Proj}_\phi: \mathbb{R}^H \to \mathbb{R}^H$ on top of its decomposed output (with a separate $\mathrm{Proj}_{\phi'}$ for the EMA target network). All baselines (Uni-modal, MMTSF~\citep{liu2024timemmd}, TaTS~\citep{li2026language}) are evaluated under identical backbone configurations to ensure a fair comparison.

\textbf{Text encoder.}
The text branch uses a frozen GPT-2~\citep{radford2019language} to encode domain-level text descriptions into $\mathbf{Z}_S$. Freezing the encoder throughout all training eliminates gradient interference from $\mathcal{L}_{\mathrm{joint}}$ and preserves pre-trained linguistic representations. The same frozen encoder is shared across all nine Time-MMD domains and both financial benchmarks, with no domain-specific fine-tuning.

\textbf{Training configuration.}
All models are optimised with Adam~\citep{kingma2015adam} using a learning rate of $10^{-4}$. We use a batch size of 32 and train for a maximum of 20 epochs with early stopping on validation MSE with a patience of 5 epochs. Training is conducted on a single NVIDIA A100 80GB GPU. All reported results are averaged over three random seeds.

\textbf{Hyperparameters.}
The EMA momentum coefficient is set to $m = 0.99$, ensuring that the target network evolves slowly enough to maintain non-trivial residual targets throughout training. The moving-average kernel width for the Trend--Noise--Event decomposition (Eq.~\ref{eq:rtrend}) defaults to 4, and the Fourier component count $K$ (Eq.~\ref{eq:rnoise}) defaults to 2. The composite loss weights are fixed at $\lambda_1 = \lambda_2 = 0.1$ across all domains and backbone configurations, with no domain-specific tuning. 

\subsection{Dataset Details}
\label{app:data}

We evaluate REST-TS on three benchmarks spanning eleven real-world domains and two financial dataset collections. Tables~\ref{tab:timemmd}, \ref{tab:fnspid}, and \ref{tab:fnf} summarise the key statistics of each dataset.

\textbf{Time-MMD.}
Time-MMD~\citep{liu2024timemmd} is the primary benchmark for multimodal time series forecasting, pairing historical multivariate time series with expert-curated natural-language reports across nine real-world domains: Agriculture, Climate, Economy, Energy, Environment, Health, Security, SocialGood, and Traffic. The domains span a wide range of temporal characteristics, from monthly macroeconomic indicators to high-frequency daily environmental measurements, making the benchmark a demanding and diverse test of multimodal forecasting. We follow the standard train/validation/test splits and evaluation protocol defined by~\citet{liu2024timemmd}.

\textbf{FNSPID.}
The Financial News and Stock Price Integration Dataset (FNSPID)~\citep{dong2024fnspid} provides daily stock price time series paired with contemporaneous news articles for six major companies spanning diverse sectors: technology (IBM, NVIDIA), finance (JPMorgan Chase), healthcare (Pfizer), automotive (Tesla), and aviation (Delta Airlines). Unlike Time-MMD, FNSPID texts are raw news articles, introducing noisier and higher-variance textual signals as a challenging out-of-domain generalisation test.

\textbf{FNF.}
The From News to Forecast (FNF) dataset~\citep{wang2024from} pairs three financial time series, Bitcoin price, Web Traffic, and Electricity Demand, with news articles at each timestamp. Together with FNSPID, FNF constitutes our out-of-domain financial generalisation benchmark for assessing whether REST-TS's residual-exclusive supervision transfers beyond the expert-curated report style of Time-MMD.

\begin{table}[t]
\centering
\caption{Overview of the numerical data in the Time-MMD datasets~\citep{liu2024timemmd}.}
\label{tab:timemmd}
\resizebox{\textwidth}{!}{%
\begin{tabular}{lccccr}
\toprule
\textbf{Dataset Name/Domain} & \textbf{Prediction Length} & \textbf{Dimension} & \textbf{Frequency} & \textbf{Number of Samples} & \textbf{Timespan} \\
\midrule
Agriculture   & \{6, 8, 10, 12\}     &  1 & Monthly &    496 & 1983 -- Present \\
Climate       & \{6, 8, 10, 12\}     &  5 & Monthly &    496 & 1983 -- Present \\
Economy       & \{6, 8, 10, 12\}     &  3 & Monthly &    423 & 1989 -- Present \\
Energy        & \{12, 24, 36, 48\}   &  9 & Weekly  &  1,479 & 1996 -- Present \\
Environment   & \{48, 96, 192, 336\} &  4 & Daily   & 11,102 & 1982 -- 2023    \\
Health        & \{12, 24, 36, 48\}   & 11 & Monthly &  1,889 & 1997 -- Present \\
Security      & \{6, 8, 10, 12\}     &  1 & Monthly &    297 & 1999 -- Present \\
Social Good   & \{6, 8, 10, 12\}     &  1 & Monthly &    900 & 1950 -- Present \\
Traffic       & \{6, 8, 10, 12\}     &  1 & Monthly &    531 & 1980 -- Present \\
\bottomrule
\end{tabular}
}
\end{table}

\begin{table}[t]
\centering
\caption{Overview of the numerical data in the FNSPID datasets~\citep{dong2024fnspid}.}
\label{tab:fnspid}
\resizebox{\textwidth}{!}{%
\begin{tabular}{lccccr}
\toprule
\textbf{Company/Stock Name} & \textbf{Prediction Length} & \textbf{Dimension} & \textbf{Frequency} & \textbf{Number of Samples} & \textbf{Timespan} \\
\midrule
Delta Airlines (DAL)   & \{6, 8, 10, 12\} & 1 & Daily & 3,020 & 2009 -- 2020 \\
IBM (IBM)              & \{6, 8, 10, 12\} & 1 & Daily & 1,258 & 2016 -- 2020 \\
JPMorgan Chase (JPM)   & \{6, 8, 10, 12\} & 1 & Daily &   755 & 2018 -- 2020 \\
NVIDIA (NVDA)          & \{6, 8, 10, 12\} & 1 & Daily & 2,516 & 2011 -- 2020 \\
Pfizer (PFE)           & \{6, 8, 10, 12\} & 1 & Daily & 1,258 & 2016 -- 2020 \\
Tesla (TSLA)           & \{6, 8, 10, 12\} & 1 & Daily &   504 & 2019 -- 2020 \\
\bottomrule
\end{tabular}
}
\end{table}

\begin{table}[t]
\centering
\caption{Overview of the numerical data in the FNF datasets~\citep{wang2024from}.}
\label{tab:fnf}
\resizebox{\textwidth}{!}{%
\begin{tabular}{lccccr}
\toprule
\textbf{Dataset Name/Domain} & \textbf{Prediction Length} & \textbf{Dimension} & \textbf{Frequency} & \textbf{Number of Samples} & \textbf{Timespan} \\
\midrule
Bitcoin Price      & \{6, 8, 10, 12\} & 1 & Daily & 4,498 & 2009 -- 2021 \\
Web Traffic        & \{6, 8, 10, 12\} & 1 & Daily &   727 & 2015 -- 2016 \\
Electricity Demand & \{6, 8, 10, 12\} & 1 & Daily & 1,461 & 2019 -- 2022 \\
\bottomrule
\end{tabular}
}
\end{table}

\subsection{Diagnostic Metrics}
\label{app:metrics}

We use two complementary metrics to diagnose text-branch behaviour throughout the paper. Both are computed from the representations produced by the text branch of each trained model, without requiring access to ground-truth labels. Together they capture two distinct aspects of text-branch quality: \emph{internal diversity} of the representations (effective rank) and \emph{external alignment} of those representations with the forecasting targets (CKA).

\paragraph{Effective rank.}
Given a matrix $\mathbf{H} \in \mathbb{R}^{N \times d}$  whose $N$ rows are text-branch output representations across $N$ inputs and whose $d$ columns are the embedding dimensions, let $\sigma_1 \geq \sigma_2 \geq \cdots \geq \sigma_d \geq 0$ be the singular values of the mean-centred $\mathbf{H}$. The effective rank~\citep{roy2007effective} is defined as:
\begin{equation}
    \mathrm{erank}(\mathbf{H})
    = \exp\!\left(-\sum_{i=1}^{d} p_i \log p_i\right),
    \qquad
    p_i = \frac{\sigma_i}{\sum_{j} \sigma_j}.
    \label{eq:erank}
\end{equation}
Intuitively, the normalised singular values $\{p_i\}$ form a probability distribution over directions of variance. The effective rank is the exponential of the Shannon entropy of this distribution, which measures how evenly variance is spread across directions. It ranges from $1$ to $d$: it equals $1$ when all variance concentrates in a single direction (i.e., all inputs are mapped to nearly the same point regardless of content, indicating text collapse), and equals $d$ when all singular values are equal (i.e., the encoder produces maximally diverse, input-dependent representations). A high effective rank therefore indicates that the text encoder is sensitive to the content of the input, which is precisely what we want to verify.

Effective rank is particularly well suited as a diagnostic for text collapse because it does not require paired labels or a reference distribution; it only requires the text representations themselves. This makes it directly applicable to any encoder regardless of architecture or training objective, allowing fair comparison across all frameworks evaluated in this paper.

\paragraph{Centred Kernel Alignment (CKA).}
While effective rank measures the internal diversity of text representations, it does not tell us whether that diversity is \emph{useful} for forecasting. A text encoder could produce diverse representations that are completely unrelated to the prediction target. To capture this complementary aspect, we use Centred Kernel Alignment (CKA)~\citep{kornblith2019similarity}, which measures whether the structure of text representations is aligned with the structure of forecasting targets.

Given text representations $\mathbf{H}_S \in \mathbb{R}^{N \times d_S}$ and forecasting target representations $\mathbf{H}_Y \in \mathbb{R}^{N \times d_Y}$ over the same $N$ samples, the linear CKA is:
\begin{equation}
    \mathrm{CKA}(\mathbf{H}_S, \mathbf{H}_Y)
    = \frac{\|\mathbf{H}_Y^\top \mathbf{H}_S\|_F^2}
           {\|\mathbf{H}_S^\top \mathbf{H}_S\|_F
            \cdot \|\mathbf{H}_Y^\top \mathbf{H}_Y\|_F}.
    \label{eq:cka}
\end{equation}
CKA equals $0$ when the two representation matrices are completely unrelated, and equals $1$ when they are identical up to an orthogonal transformation. A high CKA value indicates that samples which are similar in the text representation space are also similar in the forecasting target space, meaning the text encoder captures variation that is predictively relevant. Critically, CKA is invariant to orthogonal transformations and isotropic scaling, making it robust to differences in the dimensionality and magnitude of representations across frameworks.

Used alongside effective rank, CKA allows us to distinguish two failure modes: (i) a collapsed encoder that produces low-diversity, content-independent representations (low effective rank, low CKA), and (ii) a diverse but task-irrelevant encoder that produces varied representations that are not aligned with forecasting targets (high effective rank, low CKA). A well-functioning text branch should exhibit both high effective rank and high CKA.

\paragraph{Estimation protocol.}
For each framework-backbone pair, text-branch representations are collected from the full Time-MMD test split for a given domain. Representations are mean-centred per feature before computing both metrics. Effective rank is computed directly from the singular value decomposition of the centred representation matrix. For CKA, the forecasting target representations $\mathbf{H}_Y$ are taken as the ground-truth future values $\mathbf{Y}$ projected to the same embedding space. Both metrics are computed independently per domain and per backbone, and results are reported as radar charts across backbone architectures to enable visual comparison across frameworks.
\subsection{Algorithm}
\label{app:algorithm}
The full training algorithm is in Algorithm.~\ref{alg:rest}.

\begin{algorithm}[h]
\caption{REST-TS Training Procedure}
\label{alg:rest}
\textbf{Input:} Time series dataset $\mathcal{D} = \{\mathbf{X}, S, \mathbf{Y}\}$; prediction length $H$;
lookback window $L$; EMA momentum $m$; loss weights $\lambda_1, \lambda_2$. \\
\textbf{Output:} Trained model parameters $\Theta = \{\theta, \phi_{\mathrm{fuse}}, \phi_{\mathrm{trend}}, \phi_{\mathrm{event}}\}$.
\begin{algorithmic}[1]

\State Prepare training samples $\{\mathbf{X}^{(i)},\, S^{(i)},\, \mathbf{Y}^{(i)}\}_{i=1}^n$ of length $L$ and horizon $H$.
\State Initialise $\mathcal{F}_\theta$; EMA target $\mathcal{F}_{\theta'}$ with $\theta' \leftarrow \theta$; frozen $\mathrm{TextEnc}$; $\mathrm{Fusion}_{\phi_{\mathrm{fuse}}}$; $\mathrm{TrendHead}_{\phi_{\mathrm{trend}}}$; $\mathrm{EventHead}_{\phi_{\mathrm{event}}}$.

\While{\textit{not converged}}
    \For{\textit{each training sample} $\{\mathbf{X}^{(i)}, S^{(i)}, \mathbf{Y}^{(i)}\}$}
        \State $\mathbf{Z}_S \leftarrow \mathrm{TextEnc}(S^{(i)})$ \Comment{frozen; no gradient}
        \State $\mathbf{Z}_{\mathrm{ts}} \leftarrow \mathcal{F}_\theta(\mathbf{X}^{(i)})$;\quad $\mathbf{Z}_{\mathrm{joint}} \leftarrow \mathrm{Fusion}_{\phi_{\mathrm{fuse}}}(\mathbf{Z}_{\mathrm{ts}},\, \mathbf{Z}_S)$;\quad $\mathbf{Y}_{\mathrm{joint}} \leftarrow \mathrm{Proj}(\mathbf{Z}_{\mathrm{joint}})$
        \State $\mathbf{Y}_{\mathrm{ts}} \leftarrow \mathrm{Proj}(\mathcal{F}_{\theta'}(\mathbf{X}^{(i)}))$ \Comment{no gradient through $\theta'$};\quad $\mathbf{R} \leftarrow \mathbf{Y}^{(i)} - \mathbf{Y}_{\mathrm{ts}}$
        \State $\mathbf{R}_{\mathrm{trend}} \leftarrow \mathrm{MAvg}(\mathbf{R})$;\quad $\mathbf{R}_{\mathrm{noise}} \leftarrow \mathrm{TopK}_{\mathrm{FFT}}(\mathbf{R} - \mathbf{R}_{\mathrm{trend}},\, K)$;\quad $\mathbf{R}_{\mathrm{event}} \leftarrow \mathbf{R} - \mathbf{R}_{\mathrm{trend}} - \mathbf{R}_{\mathrm{noise}}$
        \State $\hat{\mathbf{R}}_{S,\mathrm{trend}} \leftarrow \mathrm{TrendHead}_{\phi_{\mathrm{trend}}}(\mathbf{Z}_S)$;\quad $\hat{\mathbf{R}}_{S,\mathrm{event}} \leftarrow \mathrm{EventHead}_{\phi_{\mathrm{event}}}(\mathbf{Z}_S)$
        \State $\mathcal{L} \leftarrow \mathrm{MSE}(\mathbf{Y}_{\mathrm{joint}},\mathbf{Y}^{(i)}) + \lambda_1\|\hat{\mathbf{R}}_{S,\mathrm{trend}} - \mathrm{sg}(\mathbf{R}_{\mathrm{trend}})\|_F^2 + \lambda_2\|\hat{\mathbf{R}}_{S,\mathrm{event}} - \mathrm{sg}(\mathbf{R}_{\mathrm{event}})\|_F^2$
        \State $\Theta \leftarrow \Theta - \nabla_\Theta\,\mathcal{L}$;\quad $\theta' \leftarrow m\,\theta' + (1 - m)\,\theta$
    \EndFor
\EndWhile
\State \Return $\Theta$

\end{algorithmic}
\end{algorithm}

\section{Full Results}
\label{app:full_results}

\subsection{Full Forecasting Results on TimeMMD}
\label{app:full_forecast_timemmd}

\begin{table*}[!t]
\vspace{-5pt}
\caption{Full MSE on Time-MMD using TimeBridge and iTransformer as time-series models. }
\label{tab:full_mse_tb_itrans}
\vskip 0.05in
\centering
\begin{small}
\renewcommand{\multirowsetup}{\centering}
\setlength{\tabcolsep}{3.5pt}
\resizebox{\textwidth}{!}{
\begin{tabular}{c c| c c c c c| c c c c c}
\toprule
\multicolumn{2}{c|}{\multirow{2}{*}{\textbf{Models}}} & \multicolumn{5}{c|}{TimeBridge} & \multicolumn{5}{c}{iTransformer} \\
\cmidrule(lr){3-12}
\multicolumn{2}{c|}{Method} & Uni-modal & MM-TSFLib & TaTS & REST-TS & \textit{Promotion} & Uni-modal & MM-TSFLib & TaTS & REST-TS & \textit{Promotion} \\
\toprule
\multirow{5}{*}{\scalebox{1.0}{Agriculture}} & 6 & 0.071 & 0.083 & 0.083 & 0.058 & \cellcolor{improvebg}\posc{18.3\%} & 0.077 & 0.067 & 0.066 & 0.065 & \cellcolor{improvebg}\posc{2.1\%} \\
 & 8 & 0.098 & 0.101 & 0.099 & 0.087 & \cellcolor{improvebg}\posc{11.2\%} & 0.104 & 0.101 & 0.100 & 0.098 & \cellcolor{improvebg}\posc{2.4\%} \\
 & 10 & 0.124 & 0.127 & 0.123 & 0.105 & \cellcolor{improvebg}\posc{14.7\%} & 0.142 & 0.131 & 0.124 & 0.130 & \cellcolor{worsenbd}\negc{$-$5.2\%} \\
 & 12 & 0.174 & 0.174 & 0.174 & 0.147 & \cellcolor{improvebg}\posc{15.4\%} & 0.167 & 0.178 & 0.168 & 0.158 & \cellcolor{improvebg}\posc{5.4\%} \\
\cmidrule(lr){2-12}
 & Avg & 0.117 & 0.121 & 0.120 & 0.099 & \cellcolor{improvebg}\posc{15.4\%} & 0.122 & 0.119 & 0.114 & 0.113 & \cellcolor{improvebg}\posc{1.3\%} \\
\midrule
\multirow{5}{*}{\scalebox{1.0}{Climate}} & 6 & 1.593 & 1.253 & 1.237 & 0.927 & \cellcolor{improvebg}\posc{25.1\%} & 1.127 & 1.060 & 1.048 & 0.995 & \cellcolor{improvebg}\posc{5.0\%} \\
\end{tabular}}
\end{small}
\vspace{-5pt}
\end{table*}

\begin{table*}[!t]
\vspace{-5pt}
\caption{Full MAE on Time-MMD using TimeBridge and iTransformer as time-series models.}
\label{tab:full_mae_tb_itrans}
\vskip 0.05in
\centering
\begin{small}
\renewcommand{\multirowsetup}{\centering}
\setlength{\tabcolsep}{3.5pt}
\resizebox{\textwidth}{!}{
\begin{tabular}{c c| c c c c c| c c c c c}
\toprule
\multicolumn{2}{c|}{\multirow{2}{*}{\textbf{Models}}} & \multicolumn{5}{c|}{TimeBridge} & \multicolumn{5}{c}{iTransformer} \\
\cmidrule(lr){3-12}
\multicolumn{2}{c|}{Method} & Uni-modal & MM-TSFLib & TaTS & REST-TS & \textit{Promotion} & Uni-modal & MM-TSFLib & TaTS & REST-TS & \textit{Promotion} \\
\toprule
\multirow{5}{*}{\scalebox{1.0}{Agriculture}} & 6 & 0.187 & 0.202 & 0.201 & 0.162 & \cellcolor{improvebg}\posc{13.4\%} & 0.200 & 0.180 & 0.175 & 0.173 & \cellcolor{improvebg}\posc{1.2\%} \\
 & 8 & 0.218 & 0.219 & 0.217 & 0.210 & \cellcolor{improvebg}\posc{3.2\%} & 0.228 & 0.223 & 0.224 & 0.217 & \cellcolor{improvebg}\posc{2.7\%} \\
 & 10 & 0.251 & 0.259 & 0.251 & 0.231 & \cellcolor{improvebg}\posc{7.9\%} & 0.273 & 0.257 & 0.249 & 0.250 & \cellcolor{worsenbd}\negc{$-$0.5\%} \\
 & 12 & 0.298 & 0.306 & 0.302 & 0.278 & \cellcolor{improvebg}\posc{6.7\%} & 0.301 & 0.318 & 0.309 & 0.280 & \cellcolor{improvebg}\posc{7.0\%} \\
\cmidrule(lr){2-12}
 & Avg & 0.238 & 0.246 & 0.243 & 0.220 & \cellcolor{improvebg}\posc{7.6\%} & 0.251 & 0.244 & 0.239 & 0.230 & \cellcolor{improvebg}\posc{3.8\%} \\
\midrule
\multirow{5}{*}{\scalebox{1.0}{Climate}} & 6 & 1.011 & 0.886 & 0.885 & 0.768 & \cellcolor{improvebg}\posc{13.2\%} & 0.843 & 0.820 & 0.805 & 0.784 & \cellcolor{improvebg}\posc{2.6\%} \\
\end{tabular}}
\end{small}
\vspace{-5pt}
\end{table*}

\begin{table*}[!t]
\vspace{-5pt}
\caption{Full MSE on Time-MMD using PatchTST and Crossformer as time-series models}
\label{tab:full_mse_patch_cross}
\vskip 0.05in
\centering
\begin{small}
\renewcommand{\multirowsetup}{\centering}
\setlength{\tabcolsep}{3.5pt}
\resizebox{\textwidth}{!}{
\begin{tabular}{c c| c c c c c| c c c c c}
\toprule
\multicolumn{2}{c|}{\multirow{2}{*}{\textbf{Models}}} & \multicolumn{5}{c|}{PatchTST} & \multicolumn{5}{c}{Crossformer} \\
\cmidrule(lr){3-12}
\multicolumn{2}{c|}{Method} & Uni-modal & MM-TSFLib & TaTS & REST-TS & \textit{Promotion} & Uni-modal & MM-TSFLib & TaTS & REST-TS & \textit{Promotion} \\
\toprule
\multirow{5}{*}{\scalebox{1.0}{Agriculture}} & 6 & 0.074 & 0.073 & 0.072 & 0.068 & \cellcolor{improvebg}\posc{5.5\%} & 0.222 & 0.173 & 0.330 & 0.073 & \cellcolor{improvebg}\posc{57.9\%} \\
 & 8 & 0.104 & 0.090 & 0.092 & 0.096 & \cellcolor{worsenbd}\negc{$-$6.5\%} & 0.304 & 0.242 & 0.250 & 0.099 & \cellcolor{improvebg}\posc{59.1\%} \\
 & 10 & 0.136 & 0.131 & 0.128 & 0.128 & \cellcolor{worsenbd}\negc{$-$0.3\%} & 0.357 & 0.349 & 0.336 & 0.130 & \cellcolor{improvebg}\posc{61.3\%} \\
 & 12 & 0.168 & 0.167 & 0.165 & 0.167 & \cellcolor{worsenbd}\negc{$-$0.9\%} & 0.409 & 0.442 & 0.341 & 0.160 & \cellcolor{improvebg}\posc{53.1\%} \\
\cmidrule(lr){2-12}
 & Avg & 0.120 & 0.115 & 0.114 & 0.115 & \cellcolor{worsenbd}\negc{$-$0.5\%} & 0.323 & 0.302 & 0.314 & 0.116 & \cellcolor{improvebg}\posc{61.5\%} \\
\midrule
\multirow{5}{*}{\scalebox{1.0}{Climate}} & 6 & 1.259 & 1.010 & 1.014 & 0.969 & \cellcolor{improvebg}\posc{4.1\%} & 1.159 & 0.996 & 0.949 & 0.911 & \cellcolor{improvebg}\posc{4.0\%} \\
\end{tabular}}
\end{small}
\vspace{-5pt}
\end{table*}

\begin{table*}[!t]
\vspace{-5pt}
\caption{Full MAE on Time-MMD using PatchTST and Crossformer as time-series models}
\label{tab:full_mae_patch_cross}
\vskip 0.05in
\centering
\begin{small}
\renewcommand{\multirowsetup}{\centering}
\setlength{\tabcolsep}{3.5pt}
\resizebox{\textwidth}{!}{
\begin{tabular}{c c| c c c c c| c c c c c}
\toprule
\multicolumn{2}{c|}{\multirow{2}{*}{\textbf{Models}}} & \multicolumn{5}{c|}{PatchTST} & \multicolumn{5}{c}{Crossformer} \\
\cmidrule(lr){3-12}
\multicolumn{2}{c|}{Method} & Uni-modal & MM-TSFLib & TaTS & REST-TS & \textit{Promotion} & Uni-modal & MM-TSFLib & TaTS & REST-TS & \textit{Promotion} \\
\toprule
\multirow{5}{*}{\scalebox{1.0}{Agriculture}} & 6 & 0.197 & 0.188 & 0.185 & 0.183 & \cellcolor{improvebg}\posc{1.1\%} & 0.331 & 0.321 & 0.443 & 0.182 & \cellcolor{improvebg}\posc{43.3\%} \\
 & 8 & 0.234 & 0.211 & 0.207 & 0.211 & \cellcolor{worsenbd}\negc{$-$1.9\%} & 0.406 & 0.340 & 0.347 & 0.226 & \cellcolor{improvebg}\posc{33.6\%} \\
 & 10 & 0.262 & 0.263 & 0.253 & 0.252 & \cellcolor{improvebg}\posc{0.4\%} & 0.435 & 0.415 & 0.415 & 0.273 & \cellcolor{improvebg}\posc{34.1\%} \\
 & 12 & 0.294 & 0.296 & 0.291 & 0.289 & \cellcolor{improvebg}\posc{0.8\%} & 0.451 & 0.479 & 0.385 & 0.295 & \cellcolor{improvebg}\posc{23.4\%} \\
\cmidrule(lr){2-12}
 & Avg & 0.247 & 0.239 & 0.234 & 0.234 & \cellcolor{improvebg}\posc{0.0\%} & 0.406 & 0.389 & 0.398 & 0.244 & \cellcolor{improvebg}\posc{37.2\%} \\
\midrule
\multirow{5}{*}{\scalebox{1.0}{Climate}} & 6 & 0.915 & 0.789 & 0.787 & 0.775 & \cellcolor{improvebg}\posc{1.5\%} & 0.852 & 0.760 & 0.751 & 0.752 & \cellcolor{worsenbd}\negc{$-$0.1\%} \\
\end{tabular}}
\end{small}
\vspace{-5pt}
\end{table*}

\begin{table*}[!t]
\vspace{-5pt}
\caption{Full MSE on Time-MMD using DLinear and FEDformer as time-series models}
\label{tab:full_mse_dlinear_fed}
\vskip 0.05in
\centering
\begin{small}
\renewcommand{\multirowsetup}{\centering}
\setlength{\tabcolsep}{3.5pt}
\resizebox{\textwidth}{!}{
\begin{tabular}{c c| c c c c c| c c c c c}
\toprule
\multicolumn{2}{c|}{\multirow{2}{*}{\textbf{Models}}} & \multicolumn{5}{c|}{DLinear} & \multicolumn{5}{c}{FEDformer} \\
\cmidrule(lr){3-12}
\multicolumn{2}{c|}{Method} & Uni-modal & MM-TSFLib & TaTS & REST-TS & \textit{Promotion} & Uni-modal & MM-TSFLib & TaTS & REST-TS & \textit{Promotion} \\
\toprule
\multirow{5}{*}{\scalebox{1.0}{Agriculture}} & 6 & 0.170 & 0.164 & 0.175 & 0.099 & \cellcolor{improvebg}\posc{39.6\%} & 0.091 & 0.093 & 0.110 & 0.089 & \cellcolor{improvebg}\posc{2.2\%} \\
 & 8 & 0.195 & 0.214 & 0.212 & 0.127 & \cellcolor{improvebg}\posc{34.9\%} & 0.127 & 0.188 & 0.149 & 0.116 & \cellcolor{improvebg}\posc{8.7\%} \\
 & 10 & 0.218 & 0.231 & 0.233 & 0.149 & \cellcolor{improvebg}\posc{31.7\%} & 0.154 & 0.134 & 0.169 & 0.141 & \cellcolor{worsenbd}\negc{$-$5.4\%} \\
 & 12 & 0.308 & 0.273 & 0.296 & 0.175 & \cellcolor{improvebg}\posc{35.9\%} & 0.180 & 0.225 & 0.225 & 0.176 & \cellcolor{improvebg}\posc{2.2\%} \\
\cmidrule(lr){2-12}
 & Avg & 0.223 & 0.221 & 0.229 & 0.138 & \cellcolor{improvebg}\posc{37.4\%} & 0.138 & 0.160 & 0.163 & 0.131 & \cellcolor{improvebg}\posc{5.1\%} \\
\midrule
\multirow{5}{*}{\scalebox{1.0}{Climate}} & 6 & 1.158 & 0.909 & 0.909 & 0.906 & \cellcolor{improvebg}\posc{0.3\%} & 1.206 & 1.031 & 0.973 & 0.920 & \cellcolor{improvebg}\posc{5.4\%} \\
\end{tabular}}
\end{small}
\vspace{-5pt}
\end{table*}

\begin{table*}[!t]
\vspace{-5pt}
\caption{Full MAE on Time-MMD using DLinear and FEDformer as time-series models}
\label{tab:full_mae_dlinear_fed}
\vskip 0.05in
\centering
\begin{small}
\renewcommand{\multirowsetup}{\centering}
\setlength{\tabcolsep}{3.5pt}
\resizebox{\textwidth}{!}{
\begin{tabular}{c c| c c c c c| c c c c c}
\toprule
\multicolumn{2}{c|}{\multirow{2}{*}{\textbf{Models}}} & \multicolumn{5}{c|}{DLinear} & \multicolumn{5}{c}{FEDformer} \\
\cmidrule(lr){3-12}
\multicolumn{2}{c|}{Method} & Uni-modal & MM-TSFLib & TaTS & REST-TS & \textit{Promotion} & Uni-modal & MM-TSFLib & TaTS & REST-TS & \textit{Promotion} \\
\toprule
\multirow{5}{*}{\scalebox{1.0}{Agriculture}} & 6 & 0.312 & 0.296 & 0.308 & 0.233 & \cellcolor{improvebg}\posc{21.3\%} & 0.239 & 0.235 & 0.263 & 0.210 & \cellcolor{improvebg}\posc{10.5\%} \\
 & 8 & 0.340 & 0.350 & 0.344 & 0.268 & \cellcolor{improvebg}\posc{21.2\%} & 0.281 & 0.364 & 0.321 & 0.244 & \cellcolor{improvebg}\posc{13.2\%} \\
 & 10 & 0.355 & 0.343 & 0.347 & 0.293 & \cellcolor{improvebg}\posc{14.7\%} & 0.313 & 0.265 & 0.332 & 0.271 & \cellcolor{worsenbd}\negc{$-$2.1\%} \\
 & 12 & 0.407 & 0.378 & 0.398 & 0.312 & \cellcolor{improvebg}\posc{17.5\%} & 0.313 & 0.394 & 0.384 & 0.314 & \cellcolor{worsenbd}\negc{$-$0.3\%} \\
\cmidrule(lr){2-12}
 & Avg & 0.354 & 0.342 & 0.350 & 0.276 & \cellcolor{improvebg}\posc{19.3\%} & 0.286 & 0.315 & 0.325 & 0.260 & \cellcolor{improvebg}\posc{9.1\%} \\
\midrule
\multirow{5}{*}{\scalebox{1.0}{Climate}} & 6 & 0.866 & 0.751 & 0.751 & 0.747 & \cellcolor{improvebg}\posc{0.5\%} & 0.909 & 0.804 & 0.784 & 0.753 & \cellcolor{improvebg}\posc{3.9\%} \\
\end{tabular}}
\end{small}
\vspace{-5pt}
\end{table*}

\begin{table*}[!t]
\vspace{-5pt}
\caption{Full MSE on Time-MMD using Autoformer and Informer as time-series models}
\label{tab:full_mse_auto_informer}
\vskip 0.05in
\centering
\begin{small}
\renewcommand{\multirowsetup}{\centering}
\setlength{\tabcolsep}{3.5pt}
\resizebox{\textwidth}{!}{
\begin{tabular}{c c| c c c c c| c c c c c}
\toprule
\multicolumn{2}{c|}{\multirow{2}{*}{\textbf{Models}}} & \multicolumn{5}{c|}{Autoformer} & \multicolumn{5}{c}{Informer} \\
\cmidrule(lr){3-12}
\multicolumn{2}{c|}{Method} & Uni-modal & MM-TSFLib & TaTS & REST-TS & \textit{Promotion} & Uni-modal & MM-TSFLib & TaTS & REST-TS & \textit{Promotion} \\
\toprule
\multirow{5}{*}{\scalebox{1.0}{Agriculture}} & 6 & 0.109 & 0.133 & 0.148 & 0.082 & \cellcolor{improvebg}\posc{24.8\%} & 0.451 & 0.427 & 0.269 & 0.068 & \cellcolor{improvebg}\posc{74.7\%} \\
 & 8 & 0.135 & 0.182 & 0.152 & 0.111 & \cellcolor{improvebg}\posc{17.8\%} & 0.569 & 0.562 & 0.259 & 0.095 & \cellcolor{improvebg}\posc{63.4\%} \\
 & 10 & 0.173 & 0.181 & 0.161 & 0.144 & \cellcolor{improvebg}\posc{10.7\%} & 0.618 & 0.480 & 0.392 & 0.118 & \cellcolor{improvebg}\posc{69.9\%} \\
 & 12 & 0.213 & 0.263 & 0.204 & 0.175 & \cellcolor{improvebg}\posc{14.4\%} & 0.756 & 0.782 & 0.403 & 0.150 & \cellcolor{improvebg}\posc{62.7\%} \\
\cmidrule(lr){2-12}
 & Avg & 0.158 & 0.190 & 0.166 & 0.128 & \cellcolor{improvebg}\posc{19.0\%} & 0.599 & 0.562 & 0.331 & 0.108 & \cellcolor{improvebg}\posc{67.3\%} \\
\midrule
\multirow{5}{*}{\scalebox{1.0}{Climate}} & 6 & 1.116 & 1.076 & 1.003 & 0.915 & \cellcolor{improvebg}\posc{8.8\%} & 1.084 & 1.160 & 0.939 & 0.926 & \cellcolor{improvebg}\posc{1.4\%} \\
\end{tabular}}
\end{small}
\vspace{-5pt}
\end{table*}

\begin{table*}[!t]
\vspace{-5pt}
\caption{Full MAE on Time-MMD using Autoformer and Informer as time-series models}
\label{tab:full_mae_auto_informer}
\vskip 0.05in
\centering
\begin{small}
\renewcommand{\multirowsetup}{\centering}
\setlength{\tabcolsep}{3.5pt}
\resizebox{\textwidth}{!}{
\begin{tabular}{c c| c c c c c| c c c c c}
\toprule
\multicolumn{2}{c|}{\multirow{2}{*}{\textbf{Models}}} & \multicolumn{5}{c|}{Autoformer} & \multicolumn{5}{c}{Informer} \\
\cmidrule(lr){3-12}
\multicolumn{2}{c|}{Method} & Uni-modal & MM-TSFLib & TaTS & REST-TS & \textit{Promotion} & Uni-modal & MM-TSFLib & TaTS & REST-TS & \textit{Promotion} \\
\toprule
\multirow{5}{*}{\scalebox{1.0}{Agriculture}} & 6 & 0.255 & 0.270 & 0.295 & 0.197 & \cellcolor{improvebg}\posc{22.7\%} & 0.557 & 0.543 & 0.375 & 0.180 & \cellcolor{improvebg}\posc{52.0\%} \\
 & 8 & 0.278 & 0.345 & 0.304 & 0.239 & \cellcolor{improvebg}\posc{14.0\%} & 0.633 & 0.619 & 0.373 & 0.219 & \cellcolor{improvebg}\posc{41.3\%} \\
 & 10 & 0.311 & 0.326 & 0.325 & 0.282 & \cellcolor{improvebg}\posc{9.3\%} & 0.633 & 0.533 & 0.471 & 0.232 & \cellcolor{improvebg}\posc{50.8\%} \\
 & 12 & 0.345 & 0.404 & 0.357 & 0.317 & \cellcolor{improvebg}\posc{8.1\%} & 0.698 & 0.733 & 0.458 & 0.290 & \cellcolor{improvebg}\posc{36.6\%} \\
\cmidrule(lr){2-12}
 & Avg & 0.297 & 0.336 & 0.320 & 0.259 & \cellcolor{improvebg}\posc{12.8\%} & 0.630 & 0.607 & 0.419 & 0.230 & \cellcolor{improvebg}\posc{45.2\%} \\
\midrule
\multirow{5}{*}{\scalebox{1.0}{Climate}} & 6 & 0.861 & 0.826 & 0.771 & 0.756 & \cellcolor{improvebg}\posc{2.0\%} & 0.829 & 0.865 & 0.758 & 0.760 & \cellcolor{worsenbd}\negc{$-$0.2\%} \\
\end{tabular}}
\end{small}
\vspace{-5pt}
\end{table*}

\begin{table*}[t]
\centering
\small
\caption{Full MSE results on FNF and FNSPID across all prediction horizons $T\in\{6,8,10,12\}$.}
\label{tab:full_mse}
\setlength{\tabcolsep}{3pt}
\resizebox{\textwidth}{!}{%
\begin{tabular}{c|l|r|rrr|rrrrrr}
\toprule
\multicolumn{3}{c}{} & \multicolumn{3}{c}{\textit{FNF}} & \multicolumn{6}{c}{\textit{FNSPID}} \\
\cmidrule(lr){4-6}\cmidrule(lr){7-12}
Model & Method & $T$ & Bitcoin & Electricity & Web Traffic & Delta & IBM & JPMorgan & NVIDIA & Pfizer & Tesla \\
\midrule
\multirow{15}{*}{\rotatebox{90}{PatchTST}} & \multirow{5}{*}{MM-TSFLib} & 6 & 0.011 & 0.193 & 0.676 & 0.031 & 0.216 & 0.165 & 0.113 & 0.074 & 1.011 \\
 &  & 8 & 0.013 & 0.197 & 0.730 & 0.041 & 0.285 & 0.177 & 0.125 & 0.089 & 1.297 \\
 &  & 10 & 0.006 & 0.197 & 0.669 & 0.019 & 0.323 & 0.184 & 0.147 & 0.113 & 1.572 \\
 &  & 12 & 0.021 & 0.192 & 0.743 & 0.021 & 0.306 & 0.213 & 0.175 & 0.143 & 1.775 \\
\cmidrule{3-12}
 & & Avg & 0.013 & 0.195 & 0.704 & 0.028 & 0.283 & 0.185 & 0.140 & 0.105 & 1.414 \\
\cmidrule{2-12}
 & \multirow{5}{*}{TaTS} & 6 & 0.005 & 0.188 & 0.666 & 0.014 & 0.134 & 0.095 & 0.056 & 0.037 & 0.617 \\
\bottomrule
\end{tabular}}
\end{table*}

\begin{table*}[t]
\centering
\small
\caption{Full MAE results on FNF and FNSPID across all prediction horizons $T\in\{6,8,10,12\}$.}
\label{tab:full_mae}
\setlength{\tabcolsep}{3pt}
\resizebox{\textwidth}{!}{%
\begin{tabular}{c|l|r|rrr|rrrrrr}
\toprule
\multicolumn{3}{c}{} & \multicolumn{3}{c}{\textit{FNF}} & \multicolumn{6}{c}{\textit{FNSPID}} \\
\cmidrule(lr){4-6}\cmidrule(lr){7-12}
Model & Method & $T$ & Bitcoin & Electricity & Web Traffic & Delta & IBM & JPMorgan & NVIDIA & Pfizer & Tesla \\
\midrule
\multirow{15}{*}{\rotatebox{90}{PatchTST}} & \multirow{5}{*}{MM-TSFLib} & 6 & 0.053 & 0.335 & 0.596 & 0.110 & 0.329 & 0.295 & 0.232 & 0.202 & 0.737 \\
 &  & 8 & 0.059 & 0.341 & 0.636 & 0.125 & 0.369 & 0.305 & 0.239 & 0.225 & 0.874 \\
 &  & 10 & 0.039 & 0.340 & 0.602 & 0.087 & 0.376 & 0.305 & 0.254 & 0.250 & 0.977 \\
 &  & 12 & 0.073 & 0.335 & 0.639 & 0.092 & 0.363 & 0.307 & 0.286 & 0.282 & 1.007 \\
\cmidrule{3-12}
 & & Avg & 0.056 & 0.338 & 0.618 & 0.103 & 0.359 & 0.303 & 0.253 & 0.240 & 0.899 \\
\cmidrule{2-12}
 & \multirow{5}{*}{TaTS} & 6 & 0.035 & 0.327 & 0.597 & 0.075 & 0.253 & 0.230 & 0.165 & 0.141 & 0.589 \\
\bottomrule
\end{tabular}}
\end{table*}

\begin{table}[t]
\centering
\small
\setlength{\tabcolsep}{4pt}
\caption{Full MSE and MAE of REST-TS vs.\ SpecTF on Agriculture, Climate, and Economy. \textbf{Bold} is the best.}
\label{tab:spectf_g1}
\begin{tabular}{l r cc cc c}
\toprule
\multirow{2}{*}{Dataset} & \multirow{2}{*}{$T$} & \multicolumn{2}{c}{SpecTF} & \multicolumn{2}{c}{REST-TS} & \multirow{2}{*}{\textit{Prom.}} \\
\cmidrule(lr){3-4}\cmidrule(lr){5-6}
 & & MSE & MAE & MSE & MAE & \\
\midrule
\multirow{5}{*}{Agriculture} & 6 & \textbf{0.064} & \textbf{0.170} & 0.068 & 0.177 & \cellcolor{worsenbd}-5.9\% \\
 & 8 & 0.084 & \textbf{0.192} & \textbf{0.083} & 0.198 & \cellcolor{improvebg}+1.3\% \\
 & 10 & 0.110 & 0.236 & \textbf{0.106} & \textbf{0.223} & \cellcolor{improvebg}+3.8\% \\
 & 12 & 0.152 & \textbf{0.275} & \textbf{0.144} & 0.293 & \cellcolor{improvebg}+5.3\% \\
\cmidrule(lr){2-7}
 & Avg & 0.103 & \textbf{0.218} & \textbf{0.100} & 0.223 & \cellcolor{improvebg}+2.3\% \\
\midrule
\multirow{5}{*}{Climate} & 6 & \textbf{0.904} & \textbf{0.748} & 0.904 & 0.752 & \cellcolor{worsenbd}-0.0\% \\
 & 8 & 0.927 & \textbf{0.756} & \textbf{0.926} & 0.758 & \cellcolor{improvebg}+0.1\% \\
 & 10 & 0.965 & 0.775 & \textbf{0.947} & \textbf{0.764} & \cellcolor{improvebg}+1.8\% \\
 & 12 & 0.960 & 0.770 & \textbf{0.952} & \textbf{0.767} & \cellcolor{improvebg}+0.8\% \\
\cmidrule(lr){2-7}
 & Avg & 0.939 & 0.762 & \textbf{0.932} & \textbf{0.760} & \cellcolor{improvebg}+0.7\% \\
\midrule
\multirow{5}{*}{Economy} & 6 & \textbf{0.0084} & \textbf{0.0770} & 0.0094 & 0.0810 & \cellcolor{worsenbd}-12.2\% \\
 & 8 & \textbf{0.0086} & \textbf{0.0785} & 0.0096 & 0.0825 & \cellcolor{worsenbd}-12.0\% \\
 & 10 & \textbf{0.0085} & \textbf{0.0779} & 0.0088 & 0.0783 & \cellcolor{worsenbd}-4.1\% \\
 & 12 & \textbf{0.0086} & \textbf{0.0784} & 0.0090 & 0.0791 & \cellcolor{worsenbd}-4.5\% \\
\cmidrule(lr){2-7}
 & Avg & \textbf{0.0085} & \textbf{0.0779} & 0.0092 & 0.0802 & \cellcolor{worsenbd}-8.2\% \\
\bottomrule
\end{tabular}
\end{table}

\begin{table}[t]
\centering
\small
\setlength{\tabcolsep}{4pt}
\caption{Full MSE and MAE of REST-TS vs.\ SpecTF on Energy, Environment, and Health. \textbf{Bold} is the best.}
\label{tab:spectf_g2}
\begin{tabular}{l r cc cc c}
\toprule
\multirow{2}{*}{Dataset} & \multirow{2}{*}{$T$} & \multicolumn{2}{c}{SpecTF} & \multicolumn{2}{c}{REST-TS} & \multirow{2}{*}{\textit{Prom.}} \\
\cmidrule(lr){3-4}\cmidrule(lr){5-6}
 & & MSE & MAE & MSE & MAE & \\
\midrule
\multirow{5}{*}{Energy} & 12 & 0.096 & 0.219 & \textbf{0.087} & \textbf{0.206} & \cellcolor{improvebg}+9.6\% \\
 & 24 & 0.199 & 0.326 & \textbf{0.185} & \textbf{0.309} & \cellcolor{improvebg}+7.2\% \\
 & 36 & 0.282 & 0.396 & \textbf{0.262} & \textbf{0.378} & \cellcolor{improvebg}+7.1\% \\
 & 48 & 0.406 & 0.493 & \textbf{0.340} & \textbf{0.448} & \cellcolor{improvebg}+16.3\% \\
\cmidrule(lr){2-7}
 & Avg & 0.246 & 0.359 & \textbf{0.218} & \textbf{0.335} & \cellcolor{improvebg}+11.1\% \\
\midrule
\multirow{5}{*}{Environment} & 48 & \textbf{0.265} & 0.370 & 0.265 & \textbf{0.370} & \cellcolor{worsenbd}-0.1\% \\
 & 96 & \textbf{0.265} & \textbf{0.368} & 0.267 & 0.373 & \cellcolor{worsenbd}-0.6\% \\
 & 192 & 0.255 & \textbf{0.365} & \textbf{0.255} & 0.369 & \cellcolor{improvebg}+0.1\% \\
 & 336 & \textbf{0.253} & \textbf{0.366} & 0.255 & 0.367 & \cellcolor{worsenbd}-0.7\% \\
\cmidrule(lr){2-7}
 & Avg & \textbf{0.260} & \textbf{0.367} & 0.260 & 0.369 & \cellcolor{improvebg}0.0\% \\
\midrule
\multirow{5}{*}{Health} & 12 & \textbf{0.985} & \textbf{0.647} & 1.006 & 0.664 & \cellcolor{worsenbd}-2.1\% \\
 & 24 & \textbf{1.256} & \textbf{0.718} & 1.292 & 0.744 & \cellcolor{worsenbd}-2.9\% \\
 & 36 & \textbf{1.395} & \textbf{0.769} & 1.408 & 0.790 & \cellcolor{worsenbd}-0.9\% \\
 & 48 & \textbf{1.470} & \textbf{0.797} & 1.476 & 0.816 & \cellcolor{worsenbd}-0.4\% \\
\cmidrule(lr){2-7}
 & Avg & \textbf{1.276} & \textbf{0.733} & 1.295 & 0.753 & \cellcolor{worsenbd}-1.5\% \\
\bottomrule
\end{tabular}
\end{table}

\begin{table}[t]
\centering
\small
\setlength{\tabcolsep}{4pt}
\caption{Full MSE and MAE of REST-TS vs.\ SpecTF on Security, SocialGood, and Traffic. \textbf{Bold} is the best.}
\label{tab:spectf_g3}
\begin{tabular}{l r cc cc c}
\toprule
\multirow{2}{*}{Dataset} & \multirow{2}{*}{$T$} & \multicolumn{2}{c}{SpecTF} & \multicolumn{2}{c}{REST-TS} & \multirow{2}{*}{\textit{Prom.}} \\
\cmidrule(lr){3-4}\cmidrule(lr){5-6}
 & & MSE & MAE & MSE & MAE & \\
\midrule
\multirow{5}{*}{Security} & 6 & 106.36 & \textbf{4.51} & \textbf{105.47} & 4.51 & \cellcolor{improvebg}+0.8\% \\
 & 8 & \textbf{106.95} & 4.77 & 107.47 & \textbf{4.66} & \cellcolor{worsenbd}-0.5\% \\
 & 10 & 109.77 & \textbf{4.62} & \textbf{108.69} & 4.76 & \cellcolor{improvebg}+1.0\% \\
 & 12 & 110.57 & 5.09 & \textbf{109.43} & \textbf{4.88} & \cellcolor{improvebg}+1.0\% \\
\cmidrule(lr){2-7}
 & Avg & 108.41 & 4.75 & \textbf{107.76} & \textbf{4.70} & \cellcolor{improvebg}+0.6\% \\
\midrule
\multirow{5}{*}{SocialGood} & 6 & 0.907 & \textbf{0.404} & \textbf{0.852} & 0.463 & \cellcolor{improvebg}+6.1\% \\
 & 8 & 0.939 & 0.421 & \textbf{0.858} & \textbf{0.406} & \cellcolor{improvebg}+8.6\% \\
 & 10 & 0.962 & \textbf{0.434} & \textbf{0.947} & 0.465 & \cellcolor{improvebg}+1.6\% \\
 & 12 & 1.040 & \textbf{0.514} & \textbf{0.994} & 0.521 & \cellcolor{improvebg}+4.5\% \\
\cmidrule(lr){2-7}
 & Avg & 0.962 & \textbf{0.443} & \textbf{0.913} & 0.464 & \cellcolor{improvebg}+5.1\% \\
\midrule
\multirow{5}{*}{Traffic} & 6 & 0.169 & \textbf{0.202} & \textbf{0.156} & 0.228 & \cellcolor{improvebg}+7.8\% \\
 & 8 & \textbf{0.169} & \textbf{0.210} & 0.170 & 0.242 & \cellcolor{worsenbd}-0.8\% \\
 & 10 & 0.173 & 0.218 & \textbf{0.163} & \textbf{0.217} & \cellcolor{improvebg}+5.9\% \\
 & 12 & \textbf{0.175} & \textbf{0.210} & 0.192 & 0.261 & \cellcolor{worsenbd}-9.8\% \\
\cmidrule(lr){2-7}
 & Avg & 0.171 & \textbf{0.210} & \textbf{0.170} & 0.237 & \cellcolor{improvebg}+0.7\% \\
\bottomrule
\end{tabular}
\end{table}

Tables~\ref{tab:full_mse_tb_itrans}--\ref{tab:full_mae_auto_informer} report the complete per-horizon MSE and MAE results across all nine Time-MMD domains and eight backbone architectures, grouped into pairs: TimeBridge and iTransformer (Tables~\ref{tab:full_mse_tb_itrans}--\ref{tab:full_mae_tb_itrans}), PatchTST and Crossformer (Tables~\ref{tab:full_mse_patch_cross}--\ref{tab:full_mae_patch_cross}), DLinear and FEDformer (Tables~\ref{tab:full_mse_dlinear_fed}--\ref{tab:full_mae_dlinear_fed}), and Autoformer and Informer (Tables~\ref{tab:full_mse_auto_informer}--\ref{tab:full_mae_auto_informer}). REST-TS achieves the lowest MSE in 62 of 72 backbone-domain configurations against TaTS. The largest gains are on Economy (up to 96.7\% with Informer) and Agriculture (up to 67.4\% with Informer), where the weaker backbones leave a large residual that the densely informative paired text can absorb. Climate shows improvements across all eight backbones. The 10 regressions are mostly on SocialGood and Traffic with the older Transformer backbones (Crossformer, FEDformer, Informer), where TaTS performs comparably. 

\subsection{Full Forecasting Results on financial benchmarks}
\label{app:full_forecast_fin}
Table~\ref{tab:full_mse} reports the complete per-horizon results on the FNF and FNSPID financial benchmarks across all prediction horizons $T \in \{6, 8, 10, 12\}$ for PatchTST and iTransformer. REST-TS achieves the best average MSE on the large majority of series, with the largest reductions on NVIDIA ($0.140 \to 0.059$ with PatchTST) and Pfizer ($0.105 \to 0.045$ with PatchTST), confirming that residual-exclusive supervision generalises to news-driven financial text.

\subsection{Full SpecTF Adaptation Results}
\label{app:full_forecast_spectf}

Tables~\ref{tab:spectf_g1}--\ref{tab:spectf_g3} report the full per-horizon MSE and MAE results comparing REST-TS against the original SpecTF across all nine Time-MMD domains. REST-TS improves over SpecTF in six of nine domains, with the largest gains in Energy and SocialGood. In Economy, SpecTF retains an edge due to its frequency-domain alignment being better suited to short-term economic indicators.

\subsection{Per-domain Diagnostic Results}
\label{app:per_domain}

\begin{figure}[t]
    \centering
    \includegraphics[width=\linewidth]{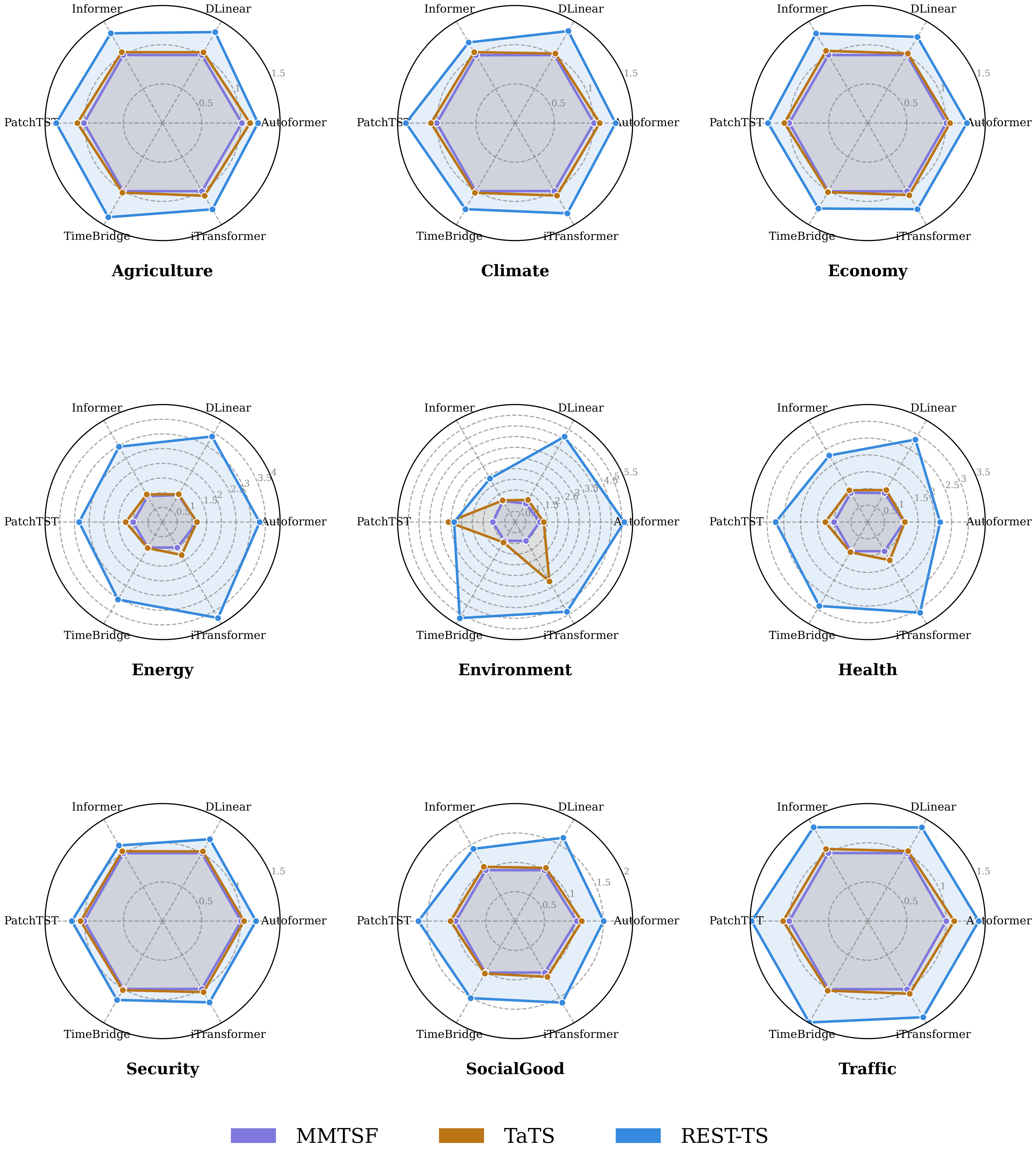}
    \caption{Effective rank of text-branch output representations ($\uparrow$) across all nine Time-MMD domains and six backbone architectures. REST-TS (blue) consistently occupies a larger area than MMTSF (purple) and TaTS (orange), with the most pronounced gaps in Energy and Environment.}
    \label{fig:all_erank}
\end{figure}

\begin{figure}[t]
    \centering
    \includegraphics[width=\linewidth]{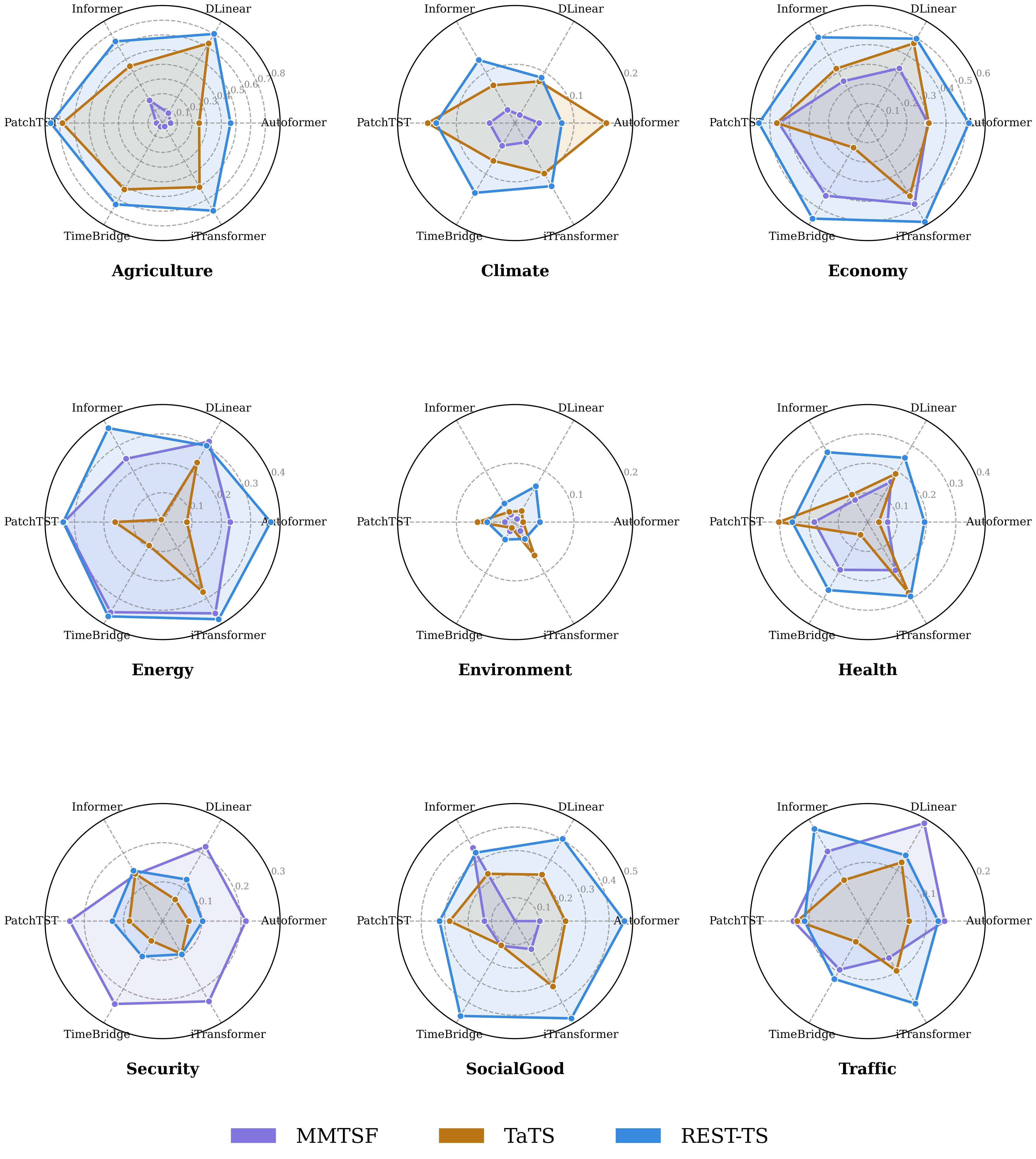}
    \caption{CKA between text-branch representations and forecasting targets ($\uparrow$) across all nine Time-MMD domains and six backbone architectures. REST-TS (blue) achieves the highest CKA in nearly every configuration. Environment is a notable exception where all frameworks score similarly low, despite REST-TS's much higher effective rank (Figure~\ref{fig:all_erank}).}
    \label{fig:all_cka}
\end{figure}

Figures~\ref{fig:all_erank} and~\ref{fig:all_cka} present the per-domain effective rank and CKA radar charts for all nine Time-MMD domains across six backbone architectures. Each axis corresponds to one backbone; a larger enclosed polygon indicates higher representational diversity (effective rank) or stronger text-to-target alignment (CKA). REST-TS (blue) consistently occupies a larger area than MMTSF (purple) and TaTS (orange) on both metrics across all nine domains, confirming that the advantage observed in the aggregated summary holds domain-by-domain and is not limited to any specific backbone.

For effective rank, the gap between REST-TS and the baselines is most pronounced in Energy and Environment, reflecting that these domains contain rich event-driven signals, policy-driven energy shocks, extreme weather events, that REST-TS's Trend-Noise-Event decomposition is specifically designed to isolate. For CKA, the advantage is consistent across most domains, with the notable exception of Environment, where all three frameworks score similarly low despite REST-TS's substantially higher effective rank.

\subsection{Comparison with modality-balancing baselines}
\label{app:ogm_comparison}
We compare REST-TS against OGM-GE~\citep{peng2022ogmge}, a representative method that adaptively rescales the gradients of the dominant modality to slow its learning relative to the weaker modality. 

These methods were primarily developed for classification, where both modalities are supervised against a shared label and imbalance manifests as one modality "winning" the gradient. In time series forecasting with paired text the issue is qualitatively different: the numerical branch can typically explain most of the target variance on its own, so the gradient available to the text branch, even after rebalancing, is small and largely noise-dominated. The text branch is not \emph{lazy} (which gradient modulation can fix) but \emph{redundant}: there is no informative residual signal directed at it. REST-TS instead supplies the text branch with a \emph{distinct} supervision target, the structured residual after numerical decomposition, so text supervision is structurally non-redundant rather than gradient-rebalanced.

Tables~\ref{tab:ogm_diagnostic}, \ref{tab:ogm_mse_avg}, and~\ref{tab:ogm_mae_avg} confirm this prediction. OGM-GE's effective rank averages 1.008 (PatchTST) and 1.044 (iTransformer) against REST-TS's 1.862 and 2.105, and REST-TS outperforms OGM-GE on every backbone-domain configuration on both MSE and MAE. The largest gaps appear on Economy ($+92.5\%$ MSE on PatchTST, $+90.3\%$ on iTransformer), where a near-constant text branch contributes nothing to event-driven residual prediction. Gradient modulation is complementary in principle but does not eliminate the structural cause of text collapse on its own.

\begin{table}[!t]
\centering
\caption{Comparison with the OGM-GE on diagnostic metrics. \textbf{Bold} is the best per row per metric.}
\label{tab:ogm_diagnostic}
\resizebox{\textwidth}{!}{
\begin{tabular}{l l cccc cccc}
\toprule
\multirow{2}{*}{Backbone} & \multirow{2}{*}{Dataset}
  & \multicolumn{4}{c}{Effective Rank}
  & \multicolumn{4}{c}{CKA} \\
\cmidrule(lr){3-6}\cmidrule(lr){7-10}
& & MMTSF & TaTS & OGM-GE & \textbf{REST-TS}
  & MMTSF & TaTS & OGM-GE & \textbf{REST-TS} \\
\midrule
\multirow{9}{*}{PatchTST} & Agriculture & 1.033 & 1.082 & 1.001 & \textbf{1.358} & 0.346 & 0.677 & 0.122 & \textbf{0.761} \\
 & Climate & 1.026 & 1.069 & 1.003 & \textbf{1.394} & 0.083 & \textbf{0.138} & 0.017 & 0.134 \\
 & Economy & 1.027 & 1.059 & 1.003 & \textbf{1.272} & 0.459 & 0.438 & 0.254 & \textbf{0.555} \\
 & Energy & 1.048 & 1.212 & 1.002 & \textbf{2.842} & \textbf{0.340} & 0.151 & 0.313 & 0.338 \\
 & Environment & 1.829 & 2.624 & 1.042 & \textbf{2.858} & 0.017 & \textbf{0.052} & 0.017 & 0.024 \\
 & Health & 1.014 & 1.220 & 1.001 & \textbf{2.741} & 0.183 & \textbf{0.277} & 0.235 & 0.257 \\
 & Security & 1.010 & 1.042 & 1.001 & \textbf{1.159} & 0.233 & 0.092 & \textbf{0.269} & 0.128 \\
 & SocialGood & 1.101 & 1.095 & 1.017 & \textbf{1.648} & 0.229 & 0.256 & 0.309 & \textbf{0.322} \\
 & Traffic & 1.017 & 1.077 & 1.001 & \textbf{1.482} & \textbf{0.130} & 0.122 & 0.079 & 0.107 \\
\cmidrule(lr){2-10}
& Avg & 1.123 & 1.276 & 1.008 & \textbf{1.862} & 0.224 & 0.245 & 0.179 & \textbf{0.292} \\
\midrule
\multirow{9}{*}{iTransformer} & Agriculture & 1.026 & 1.065 & 1.002 & \textbf{1.271} & 0.231 & 0.517 & 0.111 & \textbf{0.689} \\
 & Climate & 1.030 & 1.064 & 1.003 & \textbf{1.331} & 0.069 & 0.102 & 0.027 & \textbf{0.124} \\
 & Economy & 1.033 & 1.057 & 1.110 & \textbf{1.269} & 0.488 & 0.425 & 0.498 & \textbf{0.583} \\
 & Energy & 1.031 & 1.236 & 1.001 & \textbf{2.893} & \textbf{0.355} & 0.261 & 0.314 & 0.326 \\
 & Environment & 1.150 & 2.669 & 1.026 & \textbf{4.845} & 0.018 & \textbf{0.055} & 0.018 & 0.017 \\
 & Health & 1.010 & 1.254 & 1.002 & \textbf{3.116} & 0.185 & 0.258 & 0.238 & \textbf{0.292} \\
 & Security & 1.009 & 1.044 & 1.001 & \textbf{1.198} & 0.232 & 0.094 & \textbf{0.257} & 0.098 \\
 & SocialGood & 1.106 & 1.093 & 1.253 & \textbf{1.603} & 0.170 & 0.308 & 0.359 & \textbf{0.478} \\
 & Traffic & 1.023 & 1.068 & 1.001 & \textbf{1.417} & 0.087 & 0.102 & 0.092 & \textbf{0.162} \\
\cmidrule(lr){2-10}
& Avg & 1.046 & 1.283 & 1.044 & \textbf{2.105} & 0.204 & 0.236 & 0.213 & \textbf{0.308} \\
\bottomrule
\end{tabular}}
\end{table}

\begin{table}[!t]
\centering
\small
\caption{Average MAE comparison of REST-TS vs.\ OGM-GE on Time-MMD with PatchTST and iTransformer. \textbf{Bold} is the best per column per backbone. Results are averaged across all prediction lengths. \textit{Promotion} denotes the MAE reduction of REST-TS relative to OGM-GE.}
\label{tab:ogm_mae_avg}
\setlength{\tabcolsep}{4pt}
\resizebox{\textwidth}{!}{
\begin{tabular}{ll rrrrrrrrr}
\toprule
Backbone & Method & Agri. & Clim. & Eco. & Ener. & Envi. & Heal. & Sec. & Soc. & Traf. \\
\midrule
\multirow{3}{*}{PatchTST}
& OGM-GE & 0.305 & 0.799 & 0.246 & 0.424 & 0.426 & 0.805 & 6.275 & 0.561 & 0.342 \\
& \textbf{REST-TS} & \textbf{0.234} & \textbf{0.786} & \textbf{0.079} & \textbf{0.366} & \textbf{0.372} & \textbf{0.744} & \textbf{4.910} & \textbf{0.450} & \textbf{0.225} \\
& \textit{Promotion} & \cellcolor{improvebg}+23.3\% & \cellcolor{improvebg}+1.6\% & \cellcolor{improvebg}+67.9\% & \cellcolor{improvebg}+13.6\% & \cellcolor{improvebg}+12.6\% & \cellcolor{improvebg}+7.6\% & \cellcolor{improvebg}+21.7\% & \cellcolor{improvebg}+19.8\% & \cellcolor{improvebg}+34.3\% \\
\midrule
\multirow{3}{*}{iTransformer}
& OGM-GE & 0.309 & 0.802 & 0.235 & 0.417 & 0.404 & 0.818 & 6.420 & 0.541 & 0.483 \\
& \textbf{REST-TS} & \textbf{0.230} & \textbf{0.798} & \textbf{0.078} & \textbf{0.361} & \textbf{0.369} & \textbf{0.740} & \textbf{5.019} & \textbf{0.443} & \textbf{0.222} \\
& \textit{Promotion} & \cellcolor{improvebg}+25.5\% & \cellcolor{improvebg}+0.5\% & \cellcolor{improvebg}+66.9\% & \cellcolor{improvebg}+13.5\% & \cellcolor{improvebg}+8.7\% & \cellcolor{improvebg}+9.6\% & \cellcolor{improvebg}+21.8\% & \cellcolor{improvebg}+18.0\% & \cellcolor{improvebg}+54.0\% \\
\bottomrule
\end{tabular}}
\end{table}

\section{Full Analysis}
\label{app:full_analysis}

\subsection{Extended Gradient Analysis}
\label{app:gradient}

\begin{figure}[t]
    \centering
    \includegraphics[width=\linewidth]{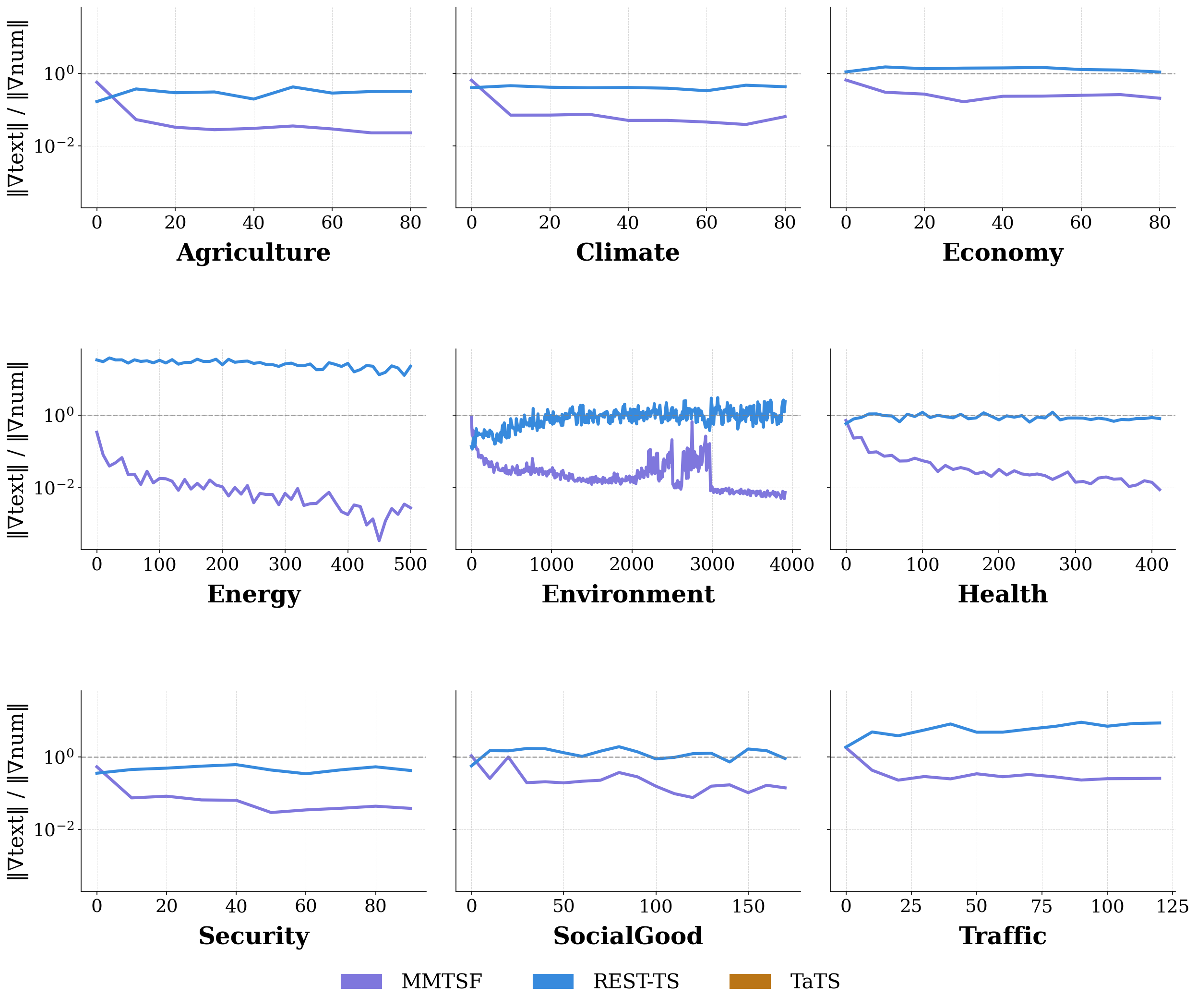}
    \caption{Text-to-numerical gradient norm ratio $\|\nabla_{\mathrm{text}}\| / \|\nabla_{\mathrm{num}}\|$ during training across all nine Time-MMD domains with iTransformer. MMTSF (purple) collapses to near-zero text gradients across every domain. REST-TS (blue) consistently maintains near-parity, demonstrating that exclusive residual supervision prevents gradient suppression regardless of domain.}
    \label{fig:grad_itransformer}
\end{figure}

\begin{figure}[t]
    \centering
    \includegraphics[width=\linewidth]{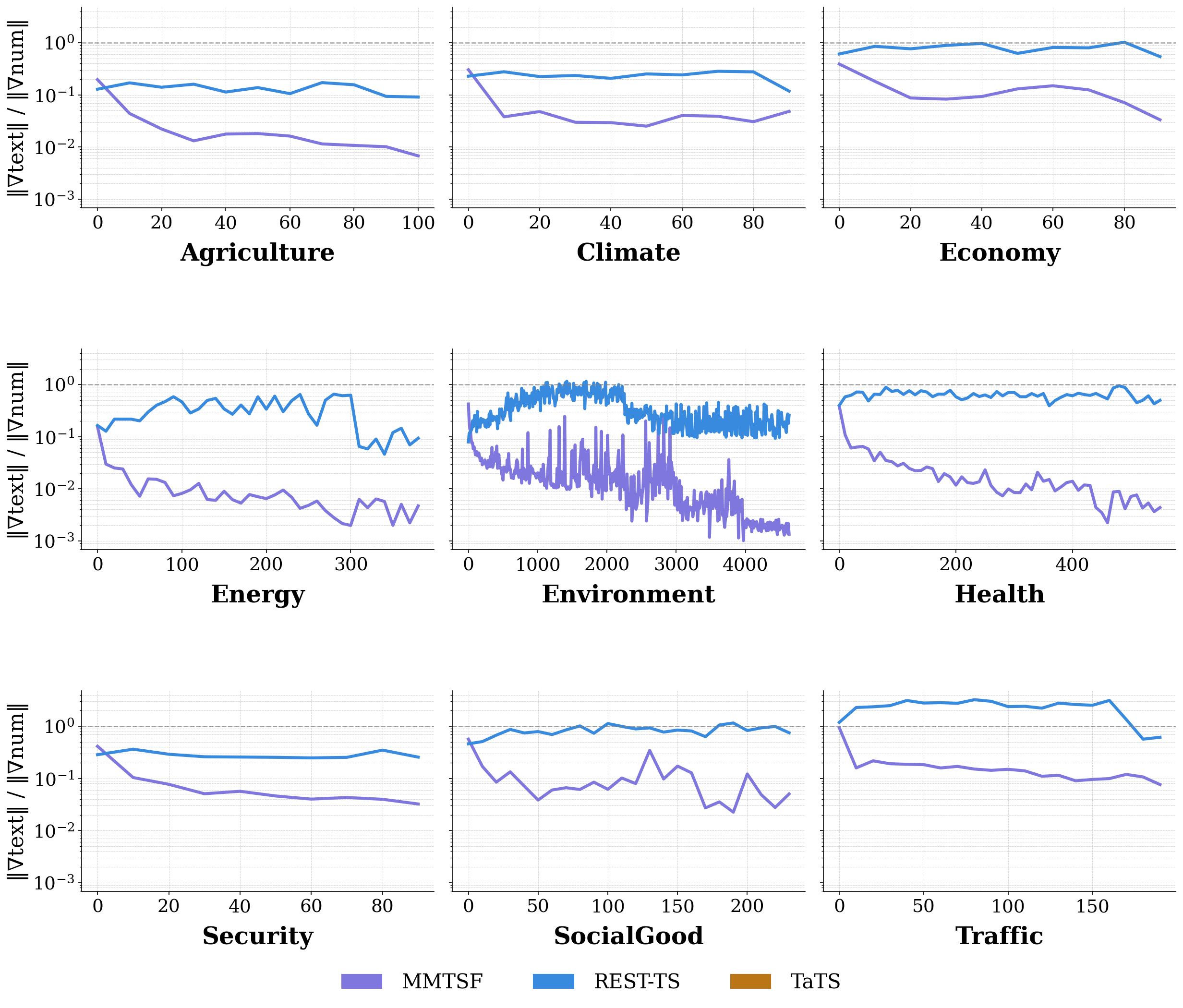}
    \caption{Text-to-numerical gradient norm ratio during training across all nine Time-MMD domains with PatchTST. The same pattern holds as with iTransformer, confirming the result is backbone-agnostic.}
    \label{fig:grad_patchtst}
\end{figure}

Figures~\ref{fig:grad_itransformer} and~\ref{fig:grad_patchtst} extend the gradient ratio analysis from Section~\ref{sec:analysis} to all nine Time-MMD domains for iTransformer and PatchTST respectively. Each subplot shows the ratio $\|\nabla_{\mathrm{text}}\| / \|\nabla_{\mathrm{num}}\|$ over training epochs, with the dashed line at $10^0$ indicating gradient parity.

The pattern is consistent across all domains and both backbones: MMTSF (purple) exhibits rapid and sustained collapse of the text gradient within early training steps, settling orders of magnitude below parity and never recovering. REST-TS (blue) maintains a gradient ratio near or above parity throughout training in every configuration. This universality reinforces that text collapse in MMTSF is a structural consequence of its fusion design, and that REST-TS's exclusive residual supervision resolves it structurally rather than through domain-specific tuning.

\subsection{Extended Learning Rate Sensitivity}
\label{app:lr_sensitivity}

\begin{figure}[t]
    \centering
    \includegraphics[width=\linewidth]{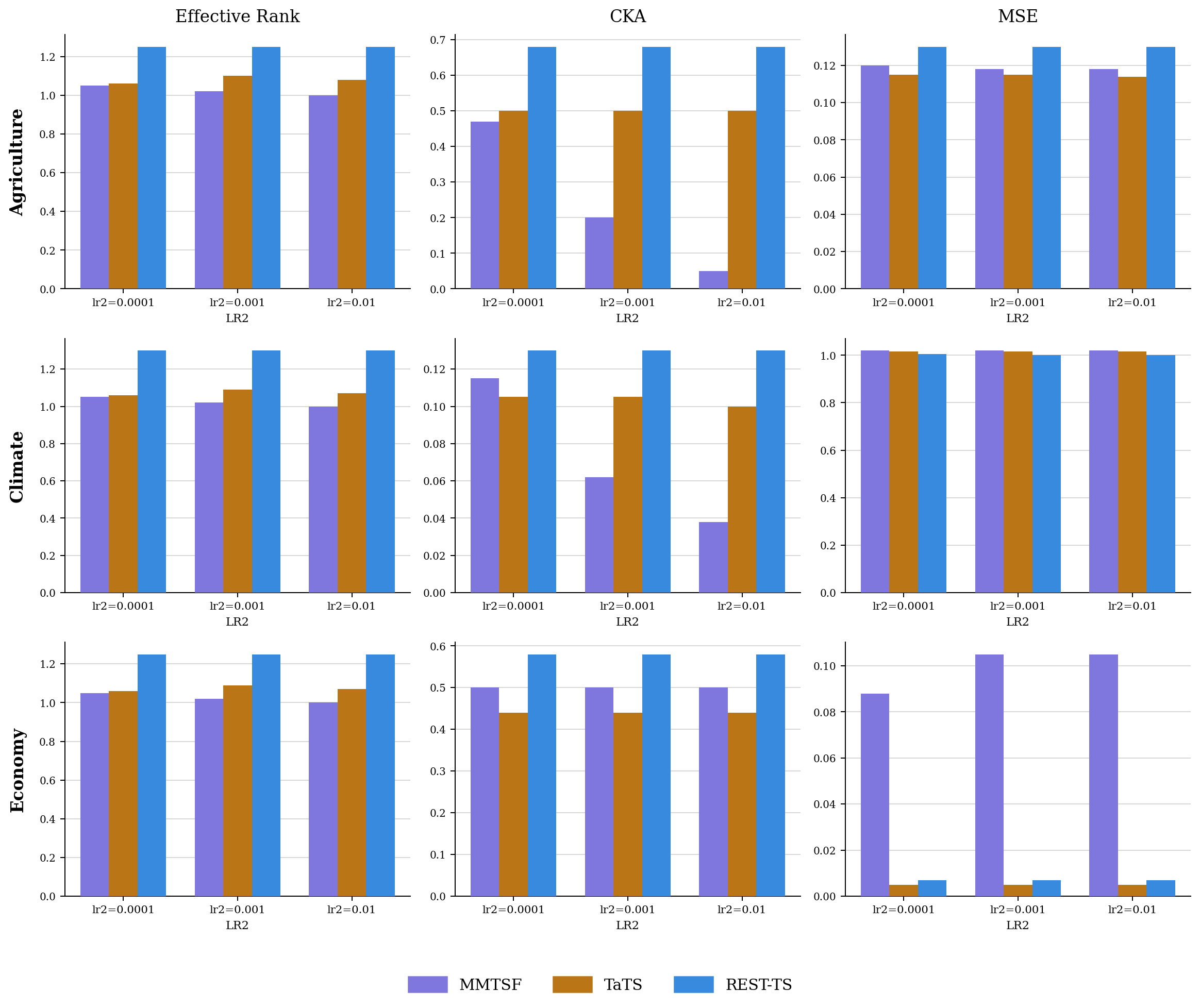}
    \caption{Sensitivity to text-branch learning rate $\mathrm{lr}_2$ for Agriculture, Climate, and Economy (iTransformer). REST-TS (blue) remains stable across all settings while MMTSF (purple) and TaTS (orange) show higher variance, particularly on forecasting error in Agriculture.}
    \label{fig:lr2_itrans_1}
\end{figure}

\begin{figure}[t]
    \centering
    \includegraphics[width=\linewidth]{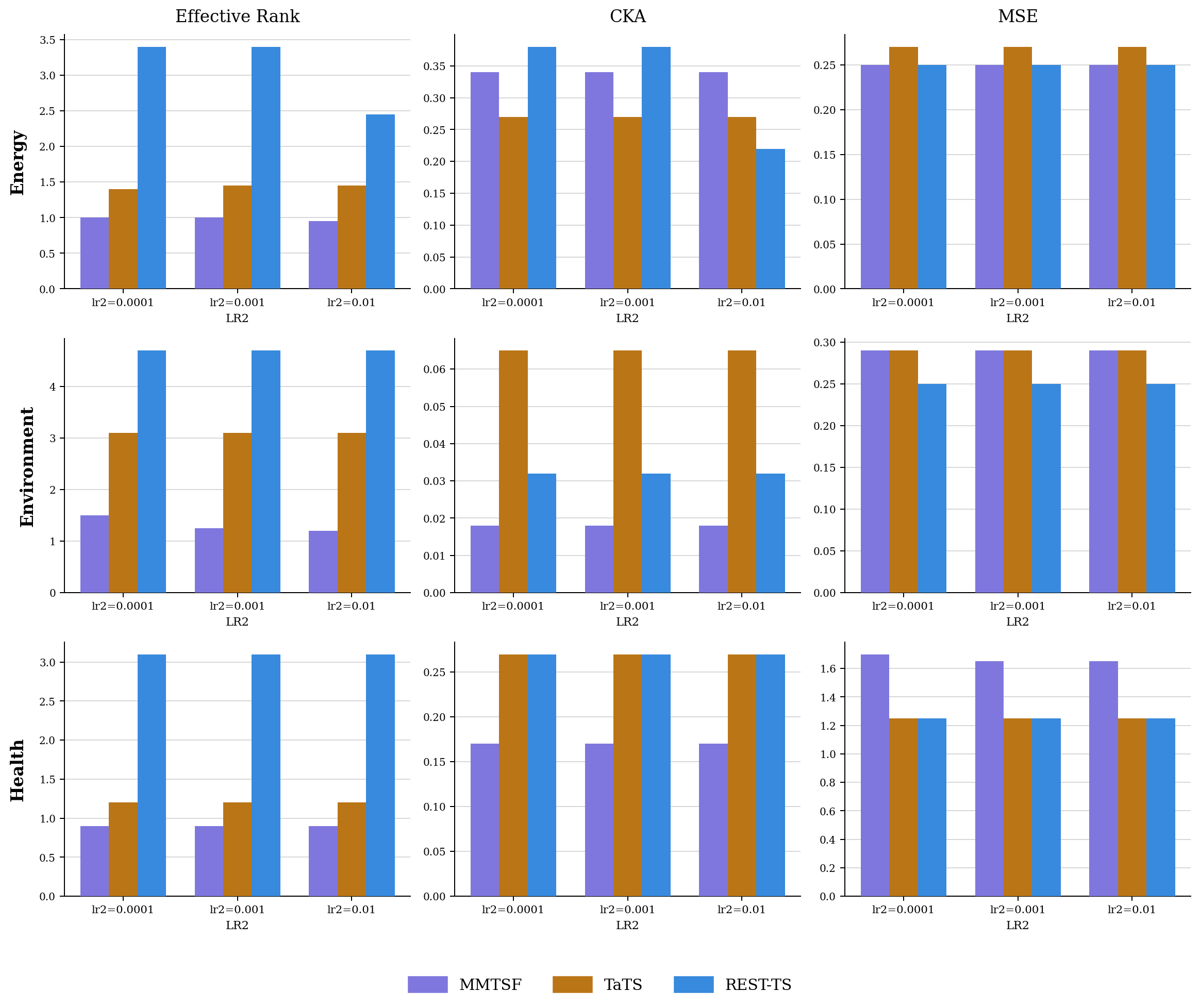}
    \caption{Sensitivity to text-branch learning rate $\mathrm{lr}_2$ for Energy, Environment, and Health (iTransformer). REST-TS (blue) is stable across all learning rates. The Energy domain shows the largest effective rank gap between REST-TS and the baselines.}
    \label{fig:lr2_itrans_2}
\end{figure}

\begin{figure}[t]
    \centering
    \includegraphics[width=\linewidth]{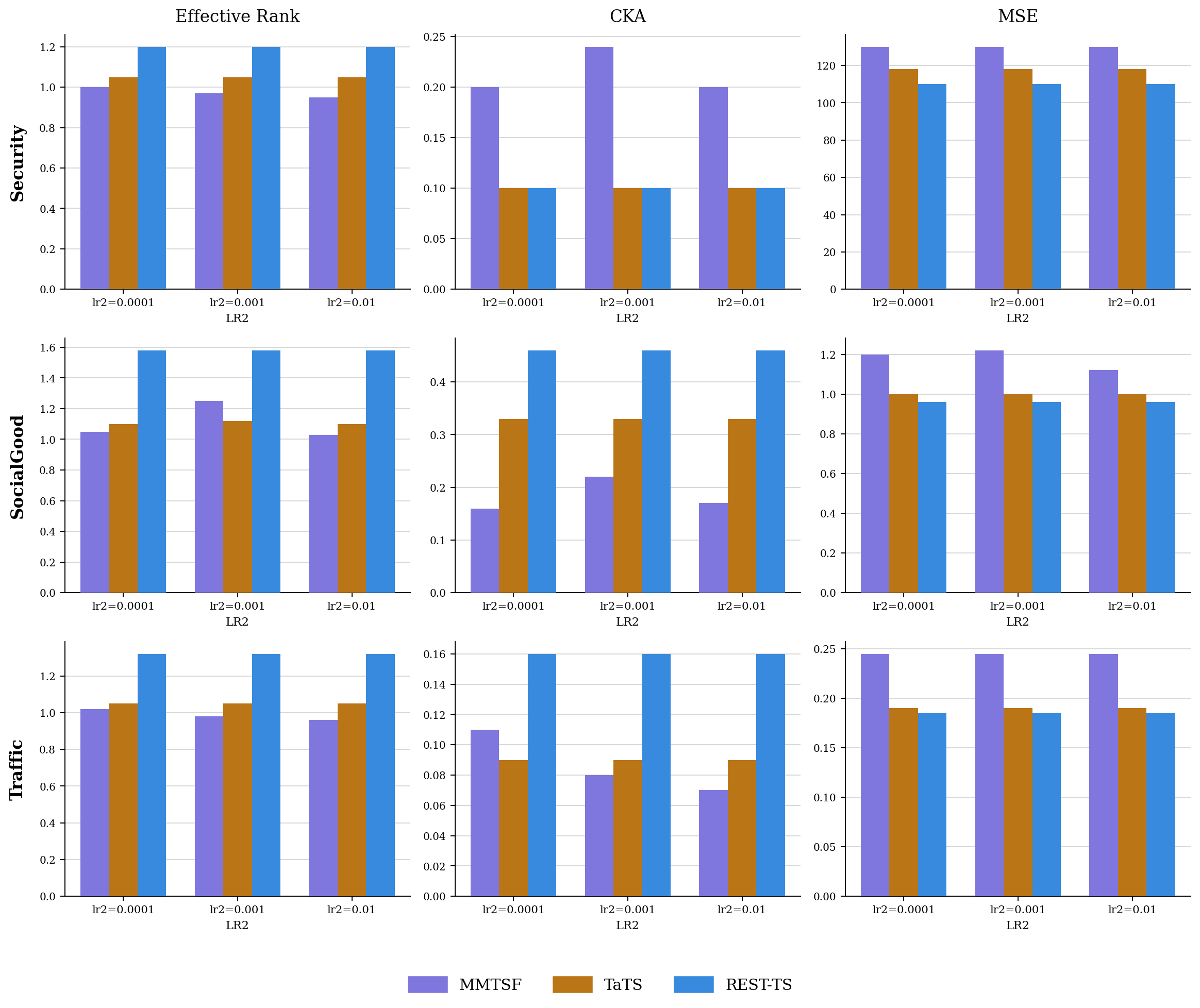}
    \caption{Sensitivity to text-branch learning rate $\mathrm{lr}_2$ for Security, SocialGood, and Traffic (iTransformer). REST-TS (blue) maintains consistent performance. MMTSF (purple) shows notable CKA degradation at higher learning rates in SocialGood.}
    \label{fig:lr2_itrans_3}
\end{figure}

Figures~\ref{fig:lr2_itrans_1}--\ref{fig:lr2_itrans_3} extend the learning rate sensitivity analysis from Section~\ref{sec:analysis} to all nine Time-MMD domains with iTransformer, evaluating effective rank, CKA, and MSE across three text-branch learning rates ($\mathrm{lr}_2 \in \{10^{-4}, 10^{-3}, 10^{-2}\}$).

REST-TS's robustness to $\mathrm{lr}_2$ holds consistently across all nine domains: REST-TS (blue) maintains near-identical effective rank, CKA, and MSE across all three learning rates, while MMTSF (purple) and TaTS (orange) exhibit higher variance particularly on forecasting error. This stability arises because REST-TS's text branch is supervised on residual decomposition targets, decoupling it from direct gradient competition with the numerical loss and making the optimal text-branch learning scale insensitive to $\mathrm{lr}_2$.

\subsection{Prediction Divergence Analysis}
\label{app:pred_divergence}

\begin{figure}[t]
    \centering
    \includegraphics[width=\linewidth]{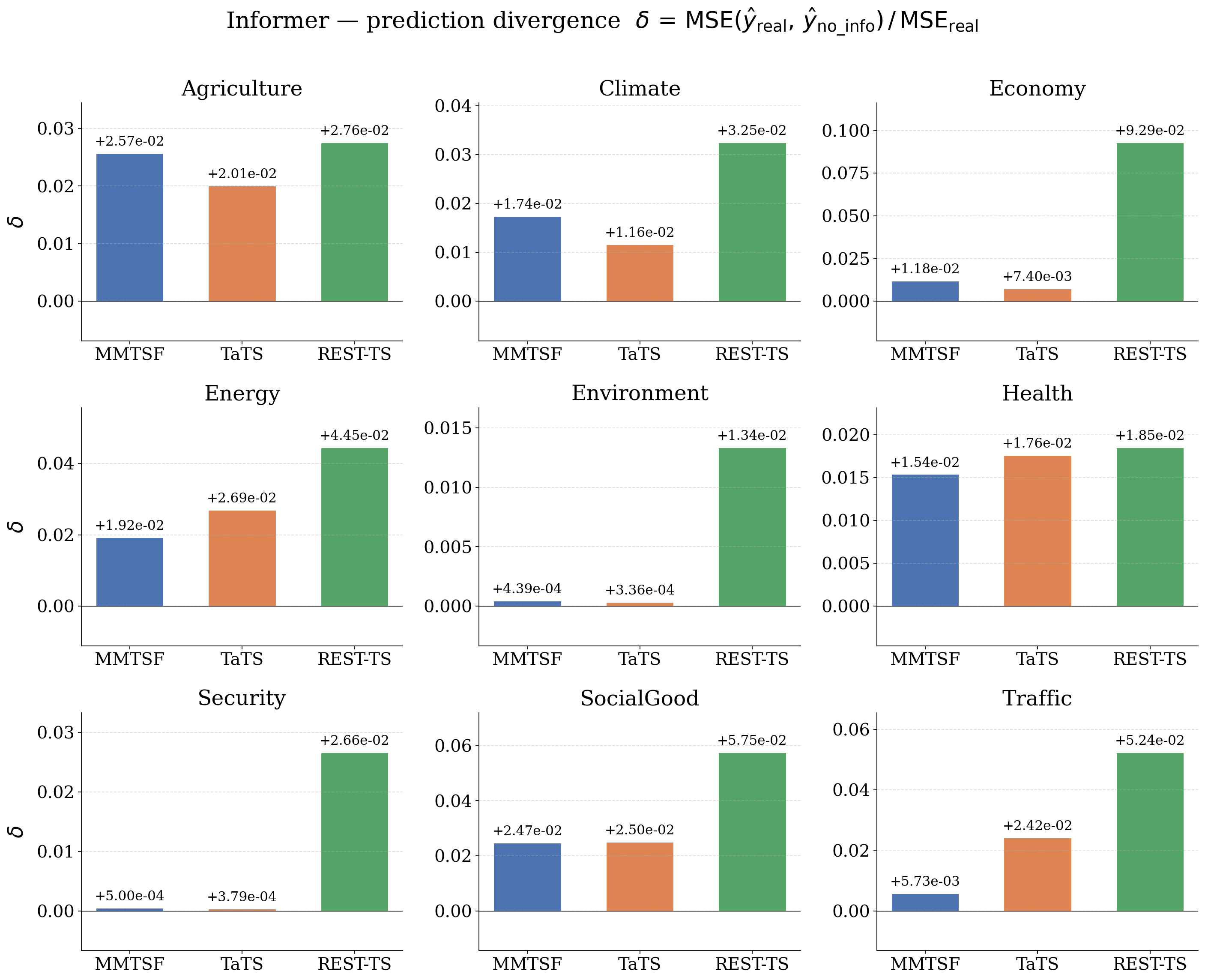}
    \caption{%
        Prediction divergence $\delta = \mathrm{MSE}(\hat{\mathbf{y}}_{\mathrm{real}},
        \hat{\mathbf{y}}_{\mathrm{no\_info}}) / \mathrm{MSE}_{\mathrm{real}}$
        across all nine Time-MMD domains with Informer. A higher $\delta$
        indicates greater sensitivity of the prediction to the text input.
        REST-TS (green) consistently achieves the highest divergence,
        while MMTSF (blue) and TaTS (orange) show near-zero values in
        several domains, confirming text collapse.
    }
    \label{fig:pred_diff_informer}
\end{figure}

\begin{figure}[t]
    \centering
    \includegraphics[width=\linewidth]{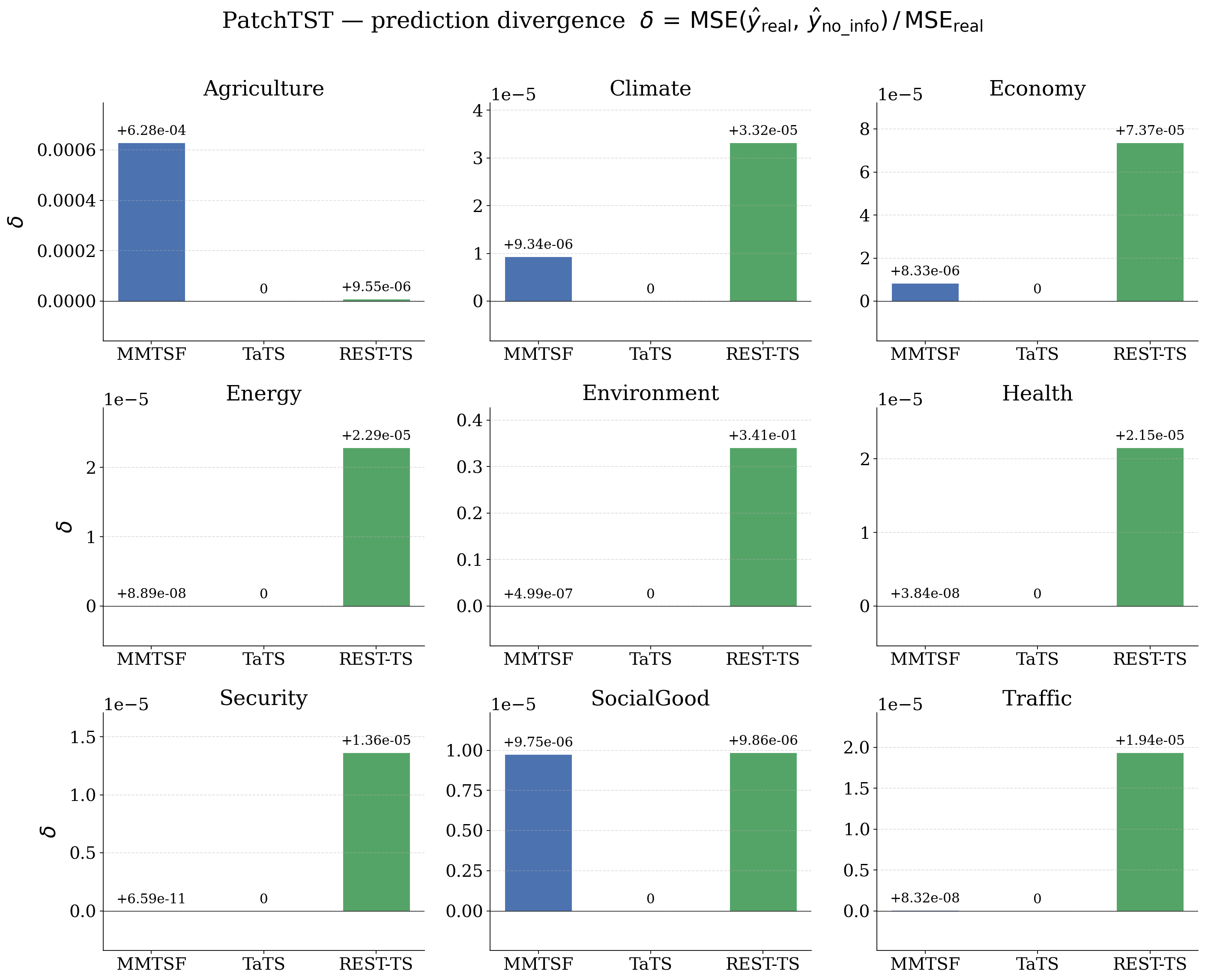}
    \caption{%
        Prediction divergence $\delta$ across all nine Time-MMD domains
        with PatchTST. TaTS (orange) yields exactly zero divergence in
        most domains, indicating that its predictions are entirely
        insensitive to the text input. REST-TS (green) maintains the
        highest divergence in every configuration, confirming that the
        result is backbone-agnostic.
    }
    \label{fig:pred_diff_patchtst}
\end{figure}

Figures~\ref{fig:pred_diff_informer} and~\ref{fig:pred_diff_patchtst} visualise the prediction divergence $\delta$ across all nine Time-MMD domains for Informer and PatchTST respectively. The metric $\delta$ is defined as:
\begin{equation}
    \delta = \frac{\mathrm{MSE}(\hat{\mathbf{y}}_{\mathrm{real}},\,
    \hat{\mathbf{y}}_{\mathrm{no\_info}})}{\mathrm{MSE}(\mathbf{y}_{\mathrm{real}}, \mathbf{y})},
\end{equation}
which measures how much the prediction changes when text information is removed, normalised by the model's own forecasting error. A higher $\delta$ indicates that the model's prediction is more sensitive to the text input, reflecting greater genuine text utilisation.

As shown in Figure~\ref{fig:pred_diff_informer}, REST-TS (green) consistently achieves the highest $\delta$ across all nine domains with Informer, with particularly large gaps over MMTSF and TaTS in event-rich domains such as Economy, Energy, and SocialGood. In contrast, MMTSF and TaTS show near-zero $\delta$ in several domains such as Security and Environment, confirming that their predictions are largely insensitive to the text input and thus suffering from text collapse. The same pattern holds for PatchTST (Figure~\ref{fig:pred_diff_patchtst}), where TaTS yields exactly zero $\delta$ across most domains, and REST-TS maintains the highest divergence in every configuration, confirming that exclusive residual supervision structurally forces the text branch to influence the final prediction.

\subsection{Hyperparameter Sensitivity}
\label{app:hp_sensitivity}

We assess REST-TS's sensitivity to the two hyperparameters of the Trend--Noise--Event decomposition: the moving-average kernel width $w$ used in Eq.~\eqref{eq:rtrend} and the Fourier component count $K$ used in Eq.~\eqref{eq:rnoise}. For each, we sweep the value across a wide range and report MAE on all nine Time-MMD domains with both Informer and TimeBridge backbones. Two patterns emerge consistently across both sweeps: the gap between the two backbones is an order of magnitude larger than the variation from either hyperparameter, and the small remaining variation is domain-dependent rather than uniform.

\begin{figure}[t]
    \centering
    \includegraphics[width=\linewidth]{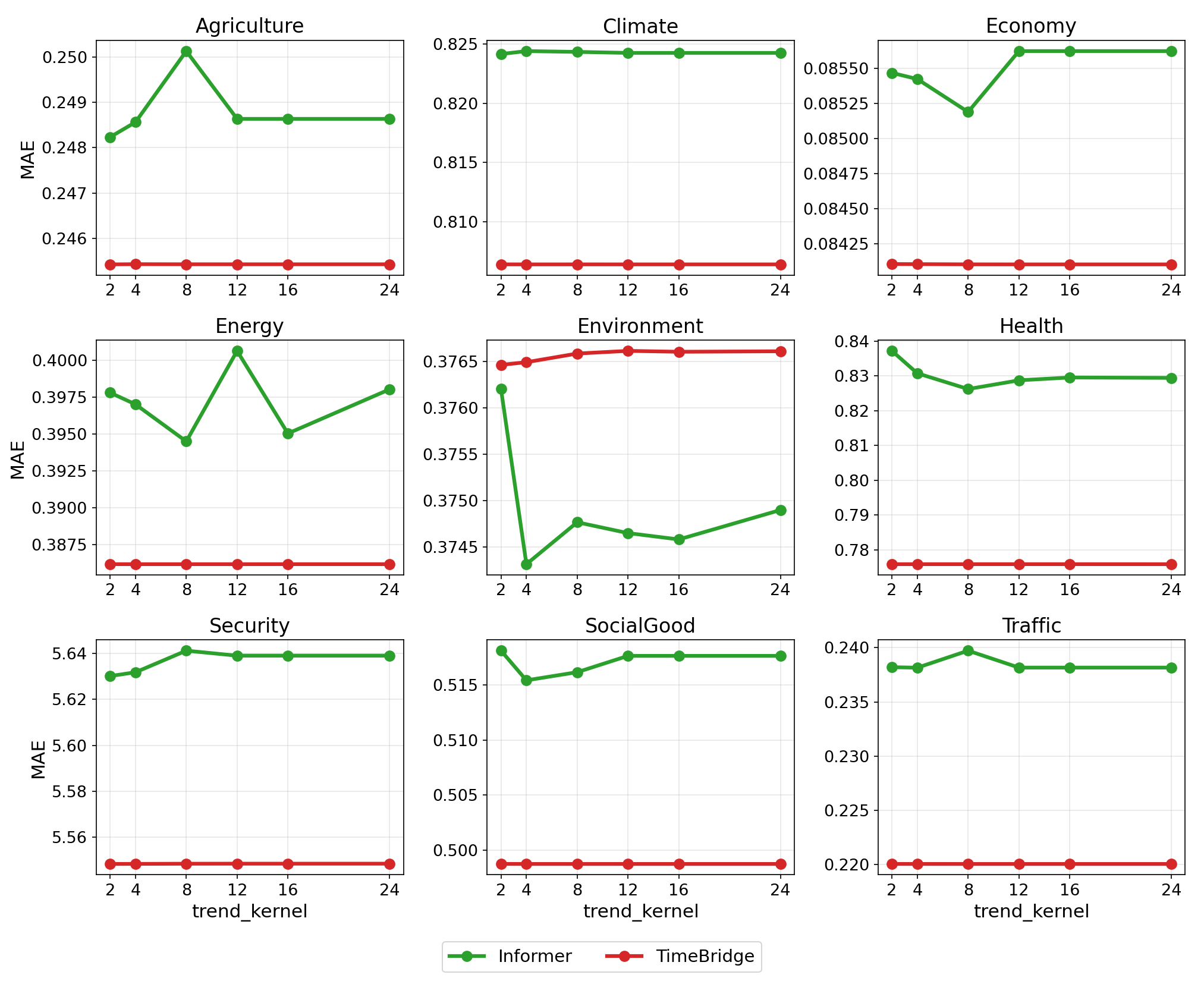}
    \caption{Sensitivity of REST-TS MAE to the moving-average trend-kernel width on all nine Time-MMD domains, for Informer (green) and TimeBridge (red) backbones.}
    \label{fig:hp_trend_kernel}
\end{figure}

\textbf{Trend kernel width.} Figure~\ref{fig:hp_trend_kernel} reports MAE as $w$ sweeps over $\{2, 4, 8, 12, 16, 24\}$. TimeBridge (red) is essentially flat on every domain, while Informer (green) wobbles non-monotonically within a band of at most a few thousandths. The wobble pattern differs across domains, with local optima at different values of $w$ on Agriculture, Energy, and Environment, and nearly flat curves on Climate, Security, and Traffic. Because no single $w$ is uniformly optimal, we interpret $w$ as a parameter to be matched to the timescale at which the trend signal in the target series varies, shorter $w$ for daily series, longer $w$ for monthly ones.

\begin{figure}[t]
    \centering
    \includegraphics[width=\linewidth]{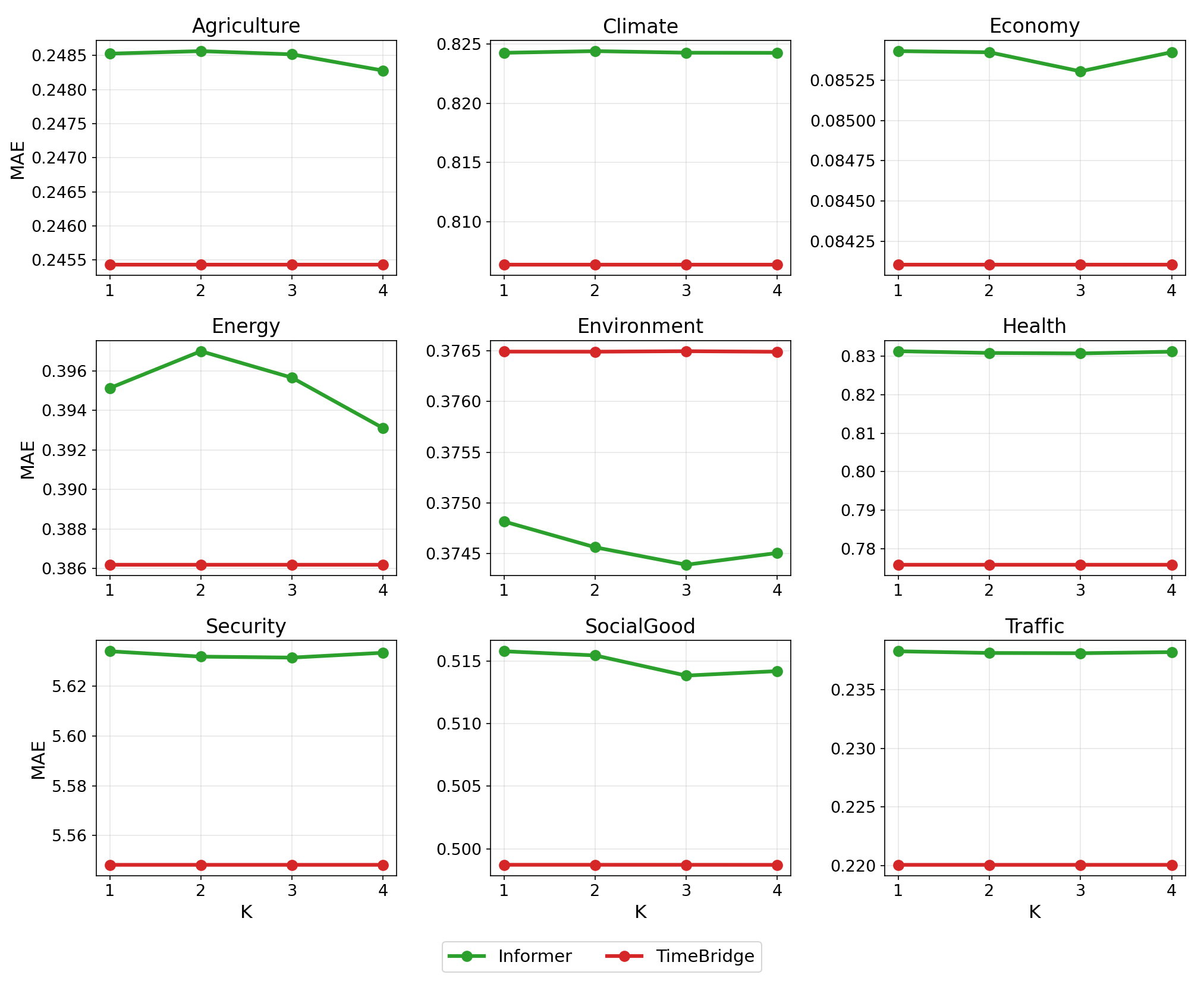}
    \caption{Sensitivity of REST-TS MAE to the Fourier component count $K$ in the noise filter on all nine Time-MMD domains, for Informer (green) and TimeBridge (red) backbones.}
    \label{fig:hp_topk}
\end{figure}

\textbf{Fourier component count.} Figure~\ref{fig:hp_topk} reports the analogous sweep over $K \in \{1, 2, 3, 4\}$. The variation is even smaller: TimeBridge is fully insensitive to $K$, and Informer's MAE varies by less than a thousandth on most domains, with the largest non-monotonic spread on Energy and SocialGood. As with $w$, no single $K$ is uniformly optimal, and we interpret $K$ as a parameter to be matched to the noise level of the target series, larger $K$ for noisier domains where high-frequency variation is uninformative, smaller $K$ for cleaner domains where preserving high-frequency content as event-attributable signal matters.

 \subsection{Prediction Visualisation}
\label{app:pred_vis}

\begin{figure}[t]
    \centering
    \includegraphics[width=\linewidth]{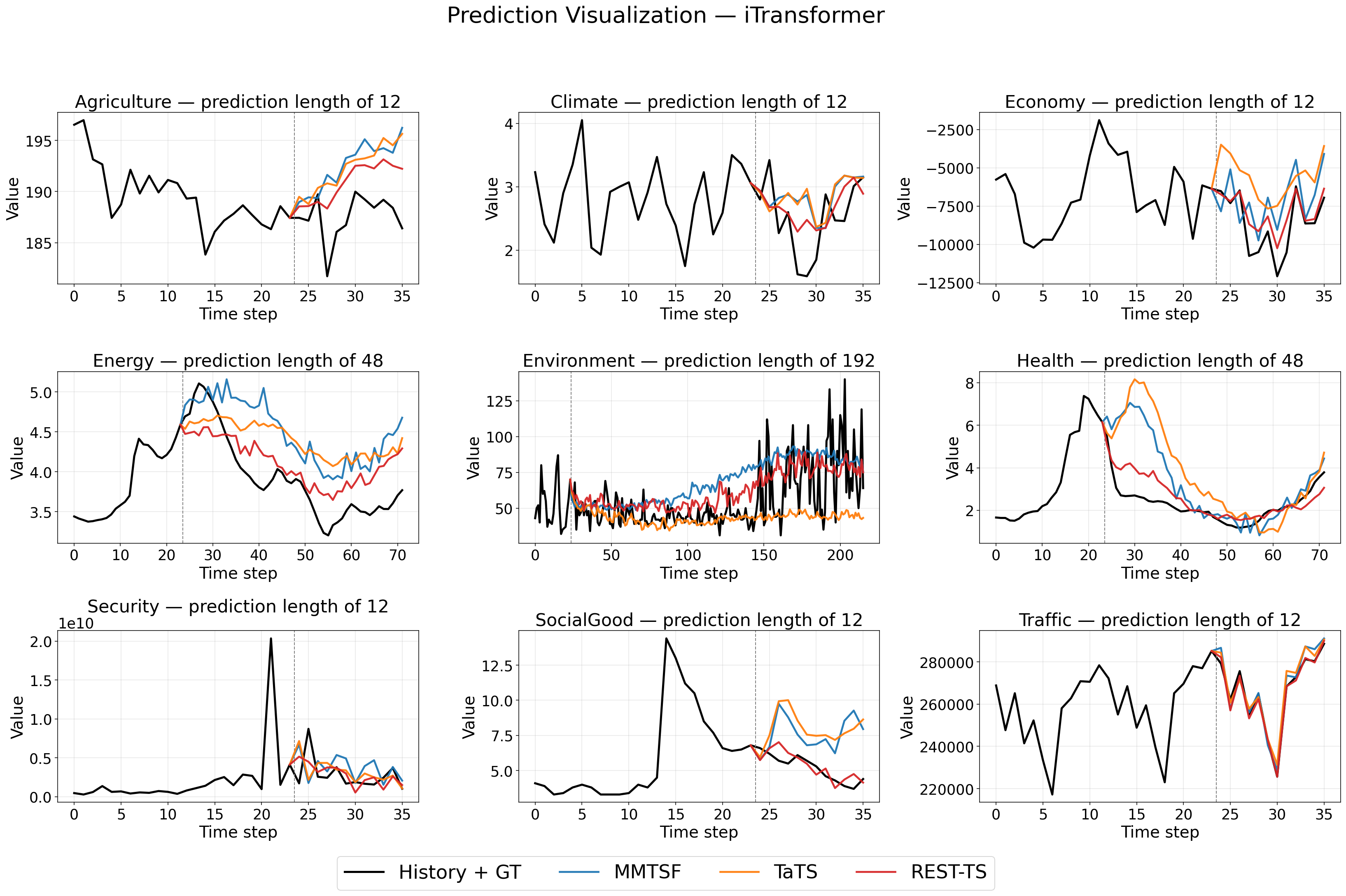}
    \caption{%
        Prediction comparison across all nine Time-MMD domains with iTransformer.
    }
    \label{fig:pred_itransformer}
\end{figure}

\begin{figure}[t]
    \centering
    \includegraphics[width=\linewidth]{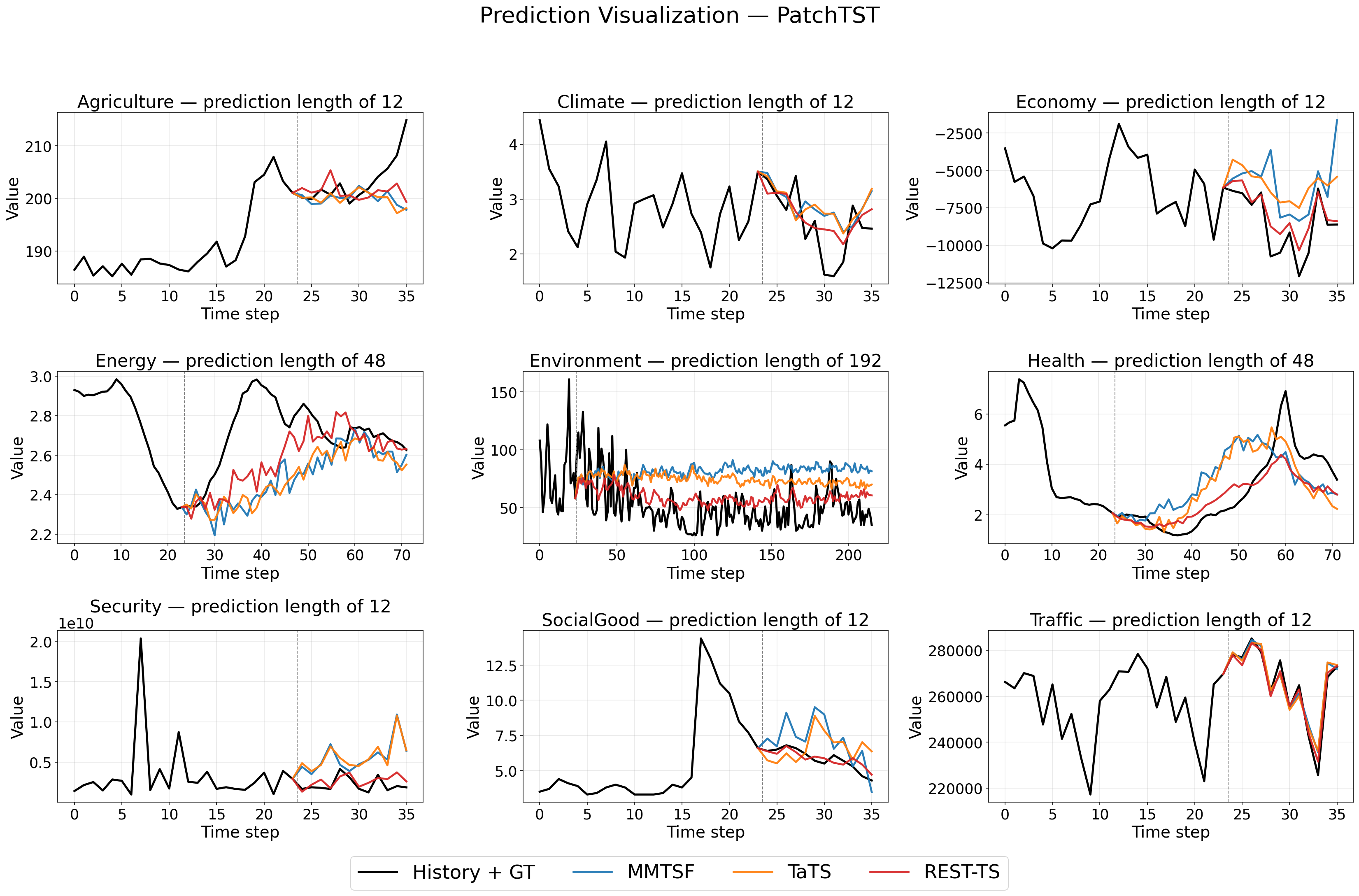}
    \caption{%
        Prediction comparison across all nine Time-MMD domains with PatchTST.
    }
    \label{fig:pred_patchtst}
\end{figure}

Figures~\ref{fig:pred_itransformer} and~\ref{fig:pred_patchtst} visualise the forecasting outputs of MMTSF, TaTS, and REST-TS across all nine Time-MMD domains for iTransformer and PatchTST respectively. Each panel shows the historical context and ground truth (black), followed by the predictions of each framework after the dashed vertical line, using the maximum prediction length of each dataset. The MSE of each framework is reported in the panel title.

REST-TS (red) consistently tracks the ground truth more closely than MMTSF (blue) and TaTS (orange) across domains. The gains are most visible in event-rich domains: in Economy, REST-TS captures the sharp directional shift in the forecast horizon that both baselines miss; in Health, REST-TS follows the steep decline more accurately; and in Security, REST-TS produces a substantially lower MSE by dampening the large spike artefacts visible in the baseline predictions. In smoother domains such as Traffic and Agriculture, all three frameworks produce similar trajectories, consistent with the modest numerical gains reported in Table~\ref{tab:main_results}.

\section{Limitations}
\label{app:limitations}
 
REST-TS has several limitations worth noting. First, the frozen GPT-2 encoder cannot adapt to highly domain-specific vocabulary, capping text-branch utility in specialised domains regardless of supervision quality. Second, the Trend--Noise--Event decomposition assumes that text-attributable variance manifests exclusively as smooth trends or discrete event spikes; domains where text describes mid-frequency or ambiguously structured phenomena may not be well served by this partition. Third, as the online backbone converges and the residual $\mathbf{R}$ shrinks, the exclusive supervision signal for the text branch weakens, which may limit content-dependent learning in near-deterministic series even with a carefully tuned EMA momentum. Finally, REST-TS assumes a single text description per forecasting window and does not handle multi-source or streaming textual inputs, and its training-time memory footprint is roughly doubled by the EMA target network, though both overheads vanish entirely at inference.

 \section{Broader Impacts}
\label{app:broader_impacts}
 
REST-TS has potential benefits across high-stakes domains covered by our evaluation — including public health, energy, and economics — where forecasting systems that genuinely incorporate textual signals such as outbreak bulletins, weather advisories, and policy reports could lead to more reliable and trustworthy predictions than existing frameworks that suffer from text collapse. The diagnostic tools we introduce (effective rank and CKA) are broadly applicable and can help the research community audit whether any multimodal model is truly leveraging its text input. On the risk side, a model that is structurally compelled to attend to text is also more sensitive to misleading or adversarially crafted inputs, and increased predictive accuracy in consequential settings warrants appropriate human oversight and uncertainty quantification. All experiments use publicly available datasets and a single NVIDIA A100 GPU; the EMA target network incurs no additional inference cost, so environmental impact is comparable to standard forecasting research.

\clearpage


\end{document}